\documentclass{article}

\usepackage[12pt]{extsizes}
\usepackage[utf8]{inputenc}
\usepackage{amsmath}
\usepackage{amssymb}
\usepackage{amsthm}
\usepackage{amsfonts}
\usepackage[numbers,sort&compress]{natbib}
\usepackage{graphicx}
\usepackage{lipsum}
\usepackage{enumitem}
\usepackage[font=footnotesize,labelfont=bf]{caption}
\usepackage{mdframed}
\usepackage{authblk}
\usepackage{longtable}
\usepackage{subfigure}
\usepackage[subfigure]{tocloft}
\usepackage[dvipsnames,table]{xcolor}
\usepackage[hidelinks]{hyperref}
\usepackage{mathtools}
\usepackage{relsize}
\usepackage{multirow}
\usepackage{booktabs}
\usepackage{tabularx}
\usepackage{bbding}
\usepackage{listings}
\usepackage{xspace}
\usepackage[ruled,vlined]{algorithm2e}
\usepackage{multirow}
\usepackage[p,osf]{cochineal}
\usepackage[scale=.95,type1]{cabin}
\usepackage[cochineal,bigdelims,cmintegrals,vvarbb]{newtxmath}
\usepackage[zerostyle=c,scaled=.94]{newtxtt}
\usepackage[cal=boondoxo]{mathalfa}
\usepackage{microtype}
\usepackage{multirow}
\usepackage{adjustbox}
\usepackage{etoolbox}
\usepackage{lmodern}
\usepackage{csquotes}

\usepackage{caption}
\usepackage{geometry}
\definecolor{myurlcolor}{HTML}{123463}
\definecolor{dc_color}{RGB}{230, 245, 244}
\definecolor{ds_color}{RGB}{195, 230, 227}
\definecolor{ms_color}{RGB}{150, 214, 209}
\hypersetup{
    colorlinks=true,
    linkcolor=black,
    filecolor=black,      
    urlcolor=black,
    citecolor=black
}

\apptocmd{\thebibliography}{\raggedright}{}{}

\makeatletter
\patchcmd{\@maketitle}{\LARGE \@title}{\fontsize{30}{19.2}\selectfont\@title}{}{}
\makeatother
\usepackage{cleveref}
\Crefname{section}{Sec.}{Secs.}
\Crefname{equation}{Eq.}{Eqs.}
\Crefname{figure}{Fig.}{Figs.}
\Crefname{tabular}{Tab.}{Tabs.}

\usepackage{amsmath,amsfonts,bm}

\def\eqref#1{equation~\ref{#1}}

\DeclareMathAlphabet{\mathsfit}{\encodingdefault}{\sfdefault}{m}{sl}
\SetMathAlphabet{\mathsfit}{bold}{\encodingdefault}{\sfdefault}{bx}{n}

\usepackage{natbib}

\usepackage{tabularx}
\usepackage{ragged2e}
\usepackage{float}

\newcolumntype{L}[1]{>{\RaggedRight\arraybackslash}p{#1}}

\title{\LARGE\textbf{Scaling Laws in Scientific Discovery with AI and Robot Scientists}}

\author{
\textbf{Pengsong Zhang}\textsuperscript{1,*}, 
\textbf{Heng Zhang}\textsuperscript{2,3,*}, 
\textbf{Huazhe Xu}\textsuperscript{4}, 
\textbf{Renjun Xu}\textsuperscript{5}, 
\textbf{Zhenting Wang}\textsuperscript{6}, 
\textbf{Cong Wang}\textsuperscript{7}, 
\textbf{Animesh Garg}\textsuperscript{8}, 
\textbf{Zhibin Li}\textsuperscript{9}, 
\textbf{Arash Ajoudani}\textsuperscript{2}, 
\textbf{Xinyu Liu}\textsuperscript{1} \\

\small *Equal contribution, \textsuperscript{1}University of Toronto, \textsuperscript{2}Istituto Italiano di Tecnologia, \textsuperscript{3}Universita di Genova, \textsuperscript{4}Tsinghua University, \textsuperscript{5}Zhejiang University, \textsuperscript{6}Rutgers University, \textsuperscript{7}Harvard University, \textsuperscript{8}Georgia Tech, \textsuperscript{9}University College of London
}

\date{}
\begin{document}
\maketitle

\begin{abstract}

Scientific discovery is poised for rapid advancement through advanced robotics and artificial intelligence. Current scientific practices face substantial limitations as manual experimentation remains time-consuming and resource-intensive, while multidisciplinary research demands knowledge integration beyond individual researchers' expertise boundaries. Here, we envision an autonomous generalist scientist (AGS) concept combines agentic AI and embodied robotics to automate the entire research lifecycle. This system could dynamically interact with both physical and virtual environments while facilitating the integration of knowledge across diverse scientific disciplines. By deploying these technologies throughout every research stage --- spanning literature review, hypothesis generation, experimentation, and manuscript writing --- and incorporating internal reflection alongside external feedback, this system aims to significantly reduce the time and resources needed for scientific discovery. Building on the evolution from virtual AI scientists to versatile generalist AI-based robot scientists, AGS promises groundbreaking potential. As these autonomous systems become increasingly integrated into the research process, we hypothesize that scientific discovery might adhere to new scaling laws, potentially shaped by the number and capabilities of these autonomous systems, offering novel perspectives on how knowledge is generated and evolves. The adaptability of embodied robots to extreme environments, paired with the flywheel effect of accumulating scientific knowledge, holds the promise of continually pushing beyond both physical and intellectual frontiers. We envision that the AGS system could catalyze a transformative shift in scientific inquiry, fostering a more efficient and innovative approach capable of overcoming current barriers, and ultimately advancing scientific progress in unprecedented ways.


\end{abstract}
\section*{Introduction}

Scientific research serves as the cornerstone of human advancement, playing a crucial role in expanding knowledge, driving technological innovation, solving complex problems, enhancing education, improving societal welfare, fostering global collaboration, stimulating economic growth, and enriching cultural and intellectual life. It not only deepens our understanding of the natural world, technological possibilities, and social phenomena but also transforms economies through the creation and refinement of technologies, ultimately elevating quality of life and productivity across societies ~\cite{gibbons1974roles, sciencenoy2023experimental}.

Despite its critical importance, the current landscape of academic research is characterized by inherent complexity and methodological constraints that frequently hinder rapid scientific advancement. Traditional research approaches necessitate labor-intensive processes, comprehensive literature analyses, and precise experimental design and execution, collectively consuming substantial time and resources ~\cite{rossoni2023barriers, vamathevan2019applications}. Furthermore, reliance on specialized expertise limits the progress and innovative capacity of research due to the dependence on a limited pool of experts ~\cite{Fabrykowska2020}. Cross-disciplinary knowledge integration serves as a pivotal factor in advancing research frontiers, particularly when addressing multifaceted global challenges in sustainable development and health sciences ~\cite{daniel2022challenges, freeth2020learning}. Multidisciplinary collaboration has yielded considerable benefits by synthesizing diverse expertise and perspectives, thus generating more comprehensive and innovative research outcomes ~\cite{toner2024artificial}. However, these collaborative efforts routinely encounter significant obstacles, including divergent disciplinary cultures ~\cite{daniel2022challenges}, specific methodologies~\cite{macleod2018makes}, and the considerable time and resources required to coordinate across fields. These persistent barriers undermine effective communication, conceptual synthesis, and the establishment of cohesive research paradigms.

Recent advancements in AI—particularly in large language models (LLMs) and foundation models ~\cite{wang2024survey}, have introduced unprecedented capabilities to generate and comprehend human-like text across multiple disciplines. Trained on vast corpora encompassing diverse fields, these models excel in applying multidisciplinary knowledge, thereby substantially enhancing scientific research ~\cite{bommasani2021opportunities,huang2024position,lu2024ai}. The intrinsic ability of generative AI to navigate and bridge disparate knowledge domains renders it exceptionally well-suited for interdisciplinary investigation ~\cite{li2024academic,mallapaty2024can}. These AI systems have exhibited remarkable proficiency in tasks ranging from information synthesis ~\cite{kang2024researcharena}, idea generation ~\cite{wang2019paperrobot, baek2024researchagent, si2024llmideas, hu2024nova}, coding ~\cite{jiang2024survey}, and academic writing ~\cite{elbanna2024exploring, lehr2024chatgpt}. Moreover, they have demonstrated autonomy in hypothesis formulation and exploration of novel scientific questions ~\cite{zenil2023future}, while also advancing specialized domains such as biomedical inquiry, image interpretation ~\cite{tu2024towardsbiomedical}, and data-driven models in medical research ~\cite{gao2024empowering}, while also fostering creativity in both scientific and artistic domains ~\cite{mediawingstrom2024redefining}. These tools not only accelerate data processing and analysis but also uncover patterns and correlations that may elude human researchers, significantly enhancing both the depth and breadth of scientific discoveries ~\cite{shir2024towards, gottweis2025towards}. However, the application of AI and LLMs remains largely confined to specific, narrow tasks or purely data-centric studies that do not involve interactions with the physical world ~\cite{lu2024ai}. This limitation highlights the need for further  development to fully realize their potential in broader scientific contexts, including more tangible, real-world applications. As the field evolves, integrating these sophisticated tools with autonomous agents and robotic systems could potentially unlock unprecedented opportunities in research and beyond~\cite{jang2024unlocking}. 
Although current LLMs still experience hallucinations, recent advancements like self-correction~\cite{kumar2024training} and recursive introspection~\cite{qu2024recursive} have gradually alleviated these concerns.

To date, the AI and robotics community have not demonstrate systems capable of integrating physical and virtual environment interactions for fully autonomous scientific research across diverse fields comparable to human scientists. A fundamental challenge is AI agent systems' limited ability to seamlessly operate across virtual and physical domains ~\cite{lu2024ai, angelopoulos2024transforming}. These systems struggle to independently access non-open scientific publications—such as those from specialized journals requiring subscription or institutional credentials, collect data that require hands-on experimentation, and performing manipulating tasks across different domains, such as laboratory procedures requiring  precise physical interaction, all essential components for conducting comprehensive research. This limitation proves especially significant in biology, medicine, and engineering, where physical world interaction is crucial. For instance, in biomedical field, AI systems should be able to handle complex physical tasks such as manipulating biological samples or operating laboratory equipments, in addition to analyzing vast amounts of virtual data ~\cite{da2024advancement}. The inability to autonomously perform these cross-domain tasks constitutes a significant barrier to developing AI scientists capable of independent research. Overcoming these challenges is critical for advancing the field and enabling AI systems to conduct scientific research with human-comparable autonomy and adaptability. Recent breakthroughs in general-purpose robotics~\cite{black2024pi_0,jiang2025brs} show promise to overcoming the limitations inherent in traditional research methodologies. These state-of-the-art robots enable seamless integration between virtual and physical experiments, thereby complementing the advancements in generative AI. By facilitating precise physical interactions—ranging from laboratory experiments to real-world manipulations—these robots not only accelerate data collection and experimentation but also enhance the reproducibility and accuracy of scientific studies. This integration marks a crucial evolution in automated research systems, paving the way for a truly autonomous research framework, thereby enhancing research productivity and broadening the horizons of academic investigation ~\cite{Open-Endednesshughesposition}.

The motivation for building autonomous scientific research system is multifaceted:

\begin{itemize}
    \item \textbf{Accelerating the Pace of Scientific Research Due to Inherent Complexity}: Contemporary scientific inquiry necessitates processing increasingly vast and multidimensional datasets that frequently exceed human cognitive capacity. Autonomous systems can systematically navigate this complexity, accelerating initial research phases and enabling researchers to advance more rapidly toward experimental validation and practical implementation.
    \item \textbf{Reducing the Demand for Specialized Expertise}: Traditional research paradigms are constrained by the requirement for highly specialized expertise, creating bottlenecks in scientific progress. Large language models can effectively synthesize and integrate knowledge across expansive document repositories, thereby enabling broader participation in generating substantive research proposals irrespective of specialized training.
    \item \textbf{Enhancing the Quality and Innovation of Research Ideas}: Developing high-quality and innovative research ideas is challenging and often requires iterative refinement. Automated systems can generate, evaluate, and systematically enhance research concepts through structured feedback loops with specialized reviewing agents, ensuring proposals meet rigorous standards of innovation and feasibility.
    \item \textbf{Promoting Cross-Disciplinary Application}: Contemporary scientific challenges increasingly transcend traditional disciplinary boundaries, requiring multidisciplinary approaches. Purpose-designed research agents with cross-domain training can collaborate synergistically, leveraging complementary expertise to address complex problems that would otherwise remain intractable to siloed research efforts.
    \item \textbf{Enhancing Reproducibility:} AI agents and robotic systems enable precise and comprehensive recording of every experimental step, from data collection to physical manipulations. This capability ensures that experiments can be reliably reproduced, addressing a major concern in scientific research~\cite{10.7554/eLife.67995,Camerer2018EvaluatingTR}.
\end{itemize}

\begin{figure}[ht]
\vskip 0.2in
\begin{center}
\centerline{\includegraphics[width=1.00\columnwidth]{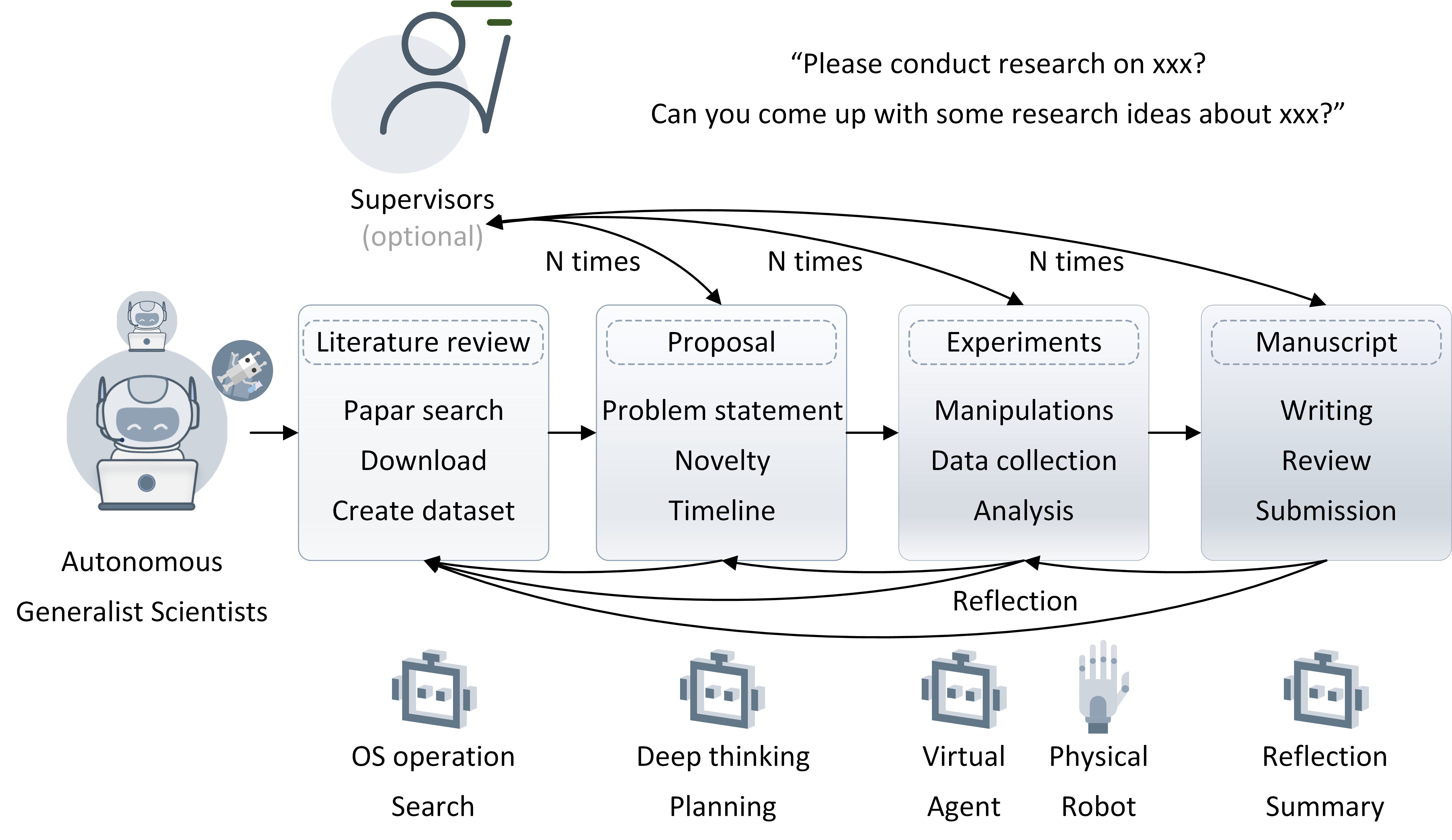}}
\caption{Framework of autonomous generalist scientist based on AI agents and robots. Research agent/robot can accelerate scientific research progress and bridge the gap between scientific knowledge in different disciplines.}
\label{ags_framework}
\end{center}
\vskip -0.2in
\end{figure}

To overcome these challenges, we envision the AGS concept, integrating agentic AI and embodied robots, equipped with universal virtual and physical manipulation abilities, capable of autonomously managing the entire research lifecycle across diverse domains. The AGS system consists of five five primary  functional modules, enhanced by integrated interaction and reflection mechanisms, as illustrated in Fig.~\ref{ags_framework}. The key modules are:

\begin{itemize}
    \item \textbf{Literature Review}: This module autonomously conducts comprehensive research analysis by simulating human-like interactions with academic databases and journal platforms. Unlike API-dependent systems, it navigates various digital environments to search, access, and manage relevant literature—even overcoming subscription barriers.

    \item \textbf{Proposal Generation}: Following literature analysis, this module formulates a comprehensive research proposal articulating a precise problem statement, well-defined objectives, and innovative hypotheses poised to advance the field. It develops detailed methodological frameworks and experimental protocols optimized for both virtual simulations and physical implementation, establishing a clear investigative roadmap.
    
    \item \textbf{Experimentation}: This module orchestrates the experimental phase of the research process, encompassing precise planning, resource optimization, and trial execution across both virtual and physical environments. Equipped with advanced robotics and AI technologies, the system performs physical manipulations, collects empirical data, and conducts virtual experiments. Furthermore, it dynamically refines experimental designs through continuous analysis of real-time results and feedback.
    
    \item \textbf{Manuscript Preparation}: Following experimental completion, this module synthesizes findings into a publication-ready manuscript. It performs comprehensive data analysis, interprets results, and formulates substantive conclusions. The system structures the document according to standard academic conventions—with methodological details, result presentations, and theoretical discussions—while conducting internal quality assessments and engaging with peer review mechanisms to ensure scholarly rigor and publication readiness.

    \item \textbf{Reflection and Feedback}: This module transcends the conventional research workflow by enabling continuous system-wide improvement. It establishes communication channels between functional components for real-time adjustments while integrating external input from human collaborators and simulated peer evaluations. Through systematic analysis of this feedback, the system refines hypotheses, methodologies, and experimental approaches, ensuring research remains responsive to emerging developments and maximizing the ultimate impact and quality of scientific outputs.

\end{itemize}

\begin{figure}[ht]
\vskip 0.2in
\begin{center}
\centerline{\includegraphics[width=1.00\columnwidth]{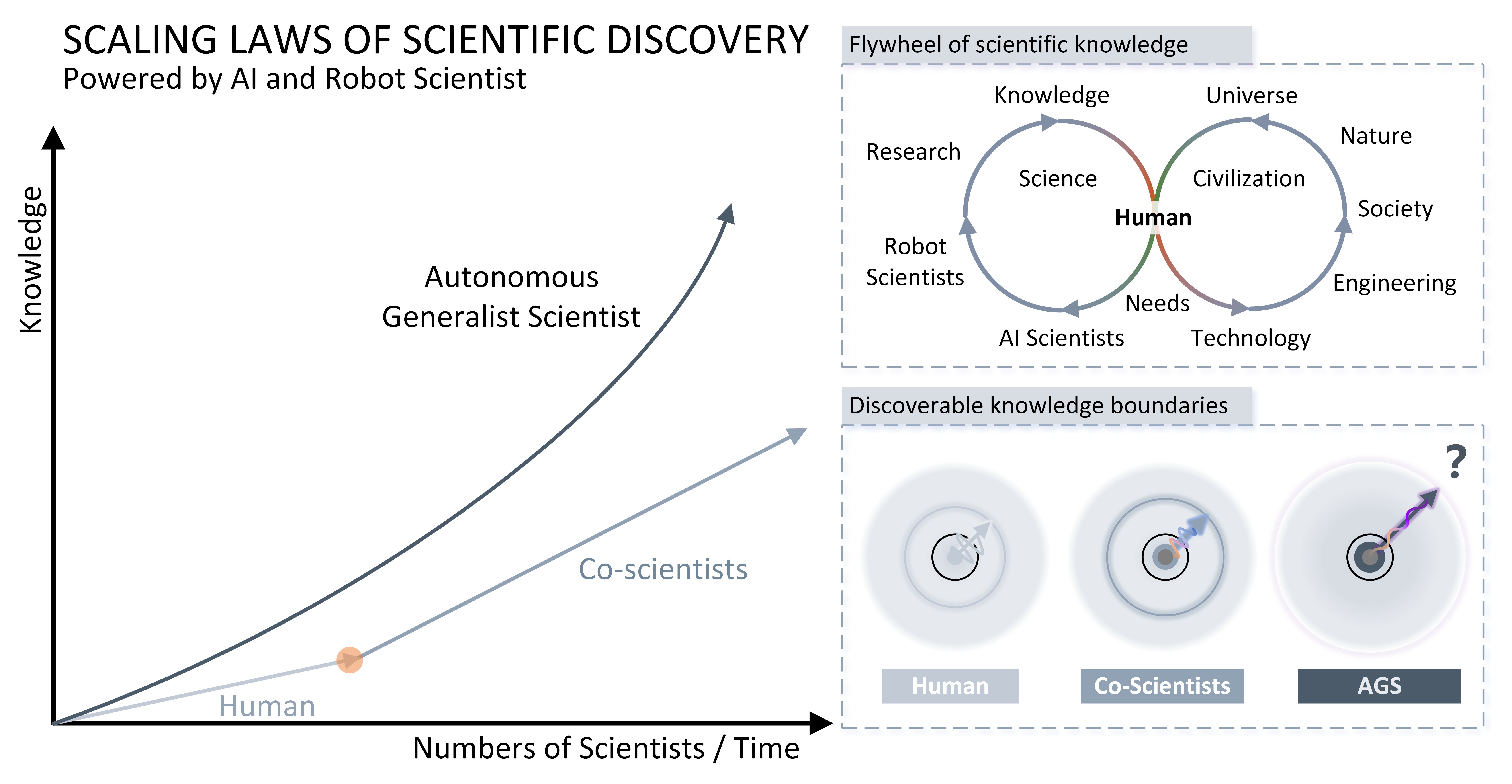}}
\caption{Evolution of scientific discovery paradigms: from human-centered research through collaborative systems to autonomous generalist scientists- breaking and transcending physical and knowledge boundaries.}
\label{ags_paradigm}
\end{center}
\vskip -0.2in
\end{figure}

Overall, the AGS represents a groundbreaking advance toward fully autonomous research systems. As shown in Fig.~\ref{ags_paradigm}, we envision the evolution of scientific research progressing from human scientists to AI and robot co-scientists, and ultimately to autonomous generalist scientists. Due to the inherent limitations in the number of human researchers, co-scientists and AGS systems will introduce new scaling laws for scientific discovery. Furthermore, the adaptability of embodied robots to extreme environments, coupled with the flywheel effect of scientific knowledge accumulation, will continuously break through both physical and knowledge boundaries. The AGS aims to pave the way for more efficient and innovative scientific investigation that transcends current limitations, ultimately accelerating the advancement of human civilization.

\section*{Definition of Automation Levels for Scientific Discovery}

\begin{figure}[ht]
\vskip 0.2in
\begin{center}
\centerline{\includegraphics[width=1.00\columnwidth]{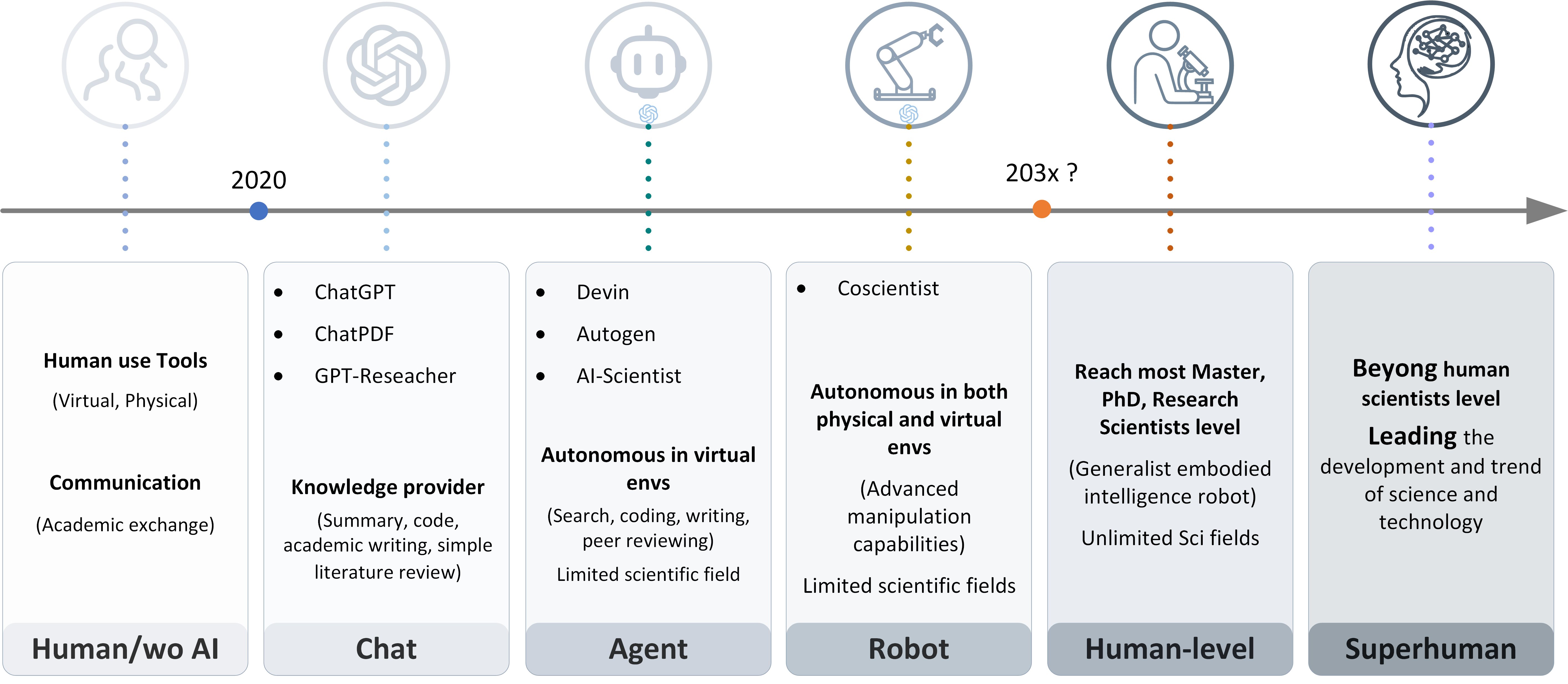}}
\caption{The timeline of automatic research with different automation levels.}
\label{ags_timeline}
\end{center}
\vskip -0.2in
\end{figure}

\begin{figure}[ht]
\vskip 0.2in
\begin{center}
\centerline{\includegraphics[width=1.00\columnwidth]{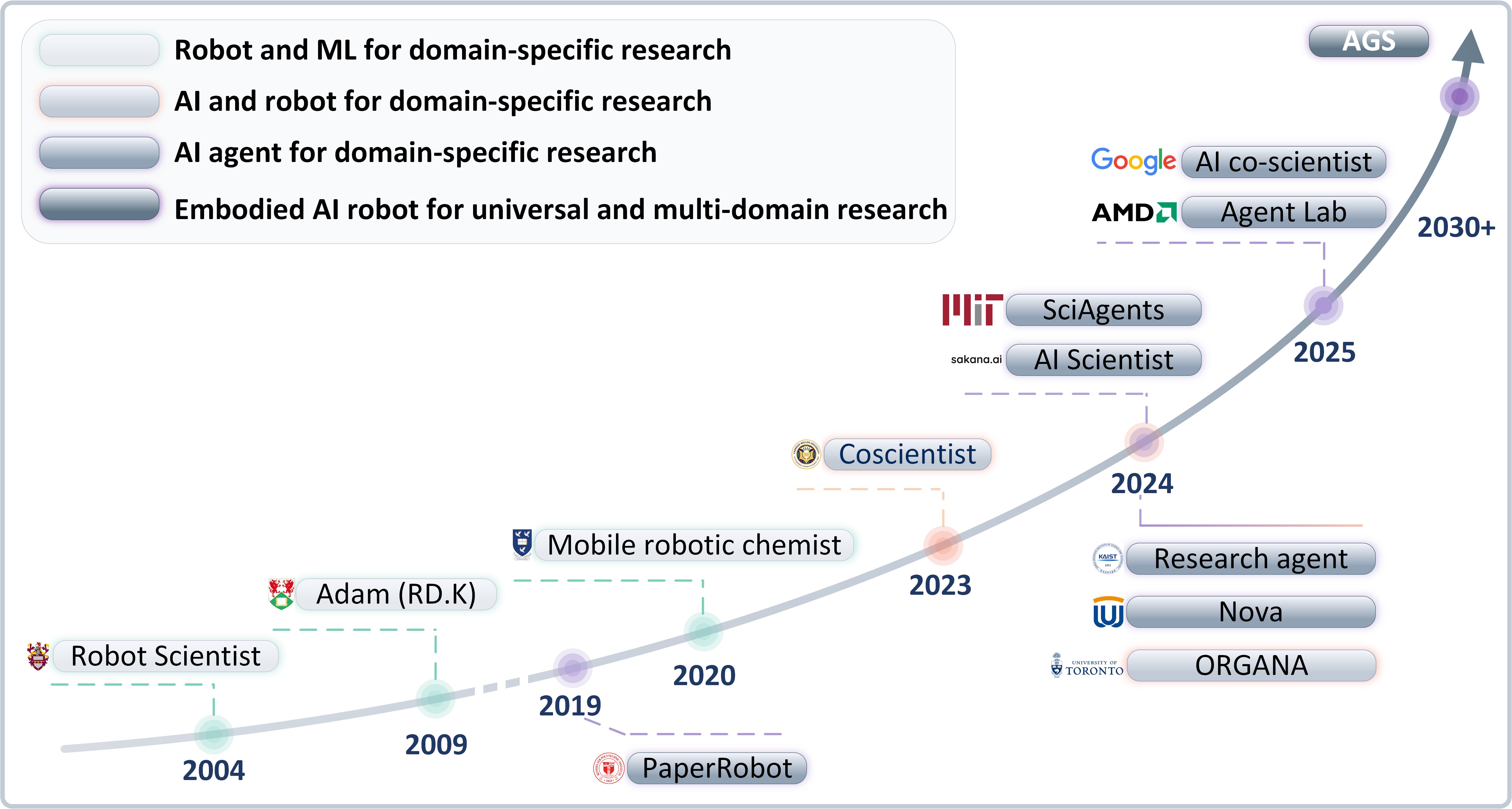}}
\caption{An overview of robot scientist evolution.}
\label{ags_evo}
\end{center}
\vskip -0.2in
\end{figure}

The AGS concept represents an AI-powered robotic system conceived to conduct research across diverse scientific domains, with the aspiration of matching and eventually exceeding the speed, scope, and depth of human scientists. This section establishes a framework for categorizing AGS into distinct levels based on their degree of autonomy, interaction with both simulated and real-world environments, and overall research capabilities (as detailed in \autoref{tab:ags_levels}). The potential evolutionary trajectory of these levels is illustrated in Figures \ref{ags_timeline} and \ref{ags_evo}.

\subsection*{Level 0: No AI}

At this foundational level, scientific inquiry is executed without the direct involvement of artificial intelligence. Research relies entirely on established methodological approaches and discipline-specific instruments. Scientists utilize specialized equipment and software tailored to particular fields—for instance, spectroscopic devices and analytical platforms in chemistry, or statistical software packages like SPSS and epidemiological modeling tools in public health. While highly effective within their designated areas, these conventional resources typically lack the capacity for seamless interdisciplinary integration and necessitate substantial human expertise for their interpretation and application.

\subsection*{Level 1: Tool-Assisted}

This level marks the introduction of simple AI tools designed to aid researchers in specific, narrowly defined tasks. Primarily driven by human scientists, the AI offers basic functionalities such as API-driven data retrieval, automated text generation, and the identification of simple connections across disciplines. Examples of systems at this level include tools like ChatGPT for text-based assistance and foundational machine learning models for data processing. While the AI can contribute by processing and summarizing information or offering suggestions in response to direct prompts, its capabilities for independent action and initiative remain limited.

\subsection*{Level 2: Intelligent Assistant}

At this stage, AI systems begin to function as sophisticated research assistants capable of navigating and synthesizing knowledge from various domains. Under human supervision, these intelligent agents can autonomously conduct web-based information gathering, perform virtual simulations, and integrate insights from diverse scientific disciplines. Systems such as OpenDevin, DeepResearch, which offer assistance in data acquisition, analysis, and the formulation of hypotheses, are representative of this level. However, significant human oversight is still required to define the scope of their activities and interpret the resulting information.

\subsection*{Level 3: Collaborative Partner}

AI systems at this level evolve into autonomous collaborative partners in scientific research, seamlessly integrating interactions with both virtual and physical environments. Equipped with advanced robotics, they can conduct experiments in domains such as biology, engineering, and medicine, performing precise manipulations in the physical world. These systems are capable of autonomously executing complex, interdisciplinary tasks but still operate in collaboration with human scientists, leveraging their respective strengths. Advanced robotic platforms that combine sensor data processing, semi-autonomous experiment execution, and integrated data analysis are key examples at this level.

\subsection*{Level 4: Autonomous Researcher}

At this stage, AI operates with a significant degree of independence, requiring only minimal human guidance. These systems possess the capacity to conduct advanced research in both simulated and real-world settings, employing autonomous information retrieval and synthesizing knowledge from a wide array of fields. They can generate novel insights and propose innovative solutions by identifying and connecting data points from previously disparate areas of study. Artificial General Intelligence Robots (AGIR) exemplify this category, pushing the boundaries of interdisciplinary research while still benefiting from occasional human oversight or intervention for complex problem-solving or ethical considerations.

\subsection*{Level 5: Pioneer}

The highest level represents fully autonomous systems that surpass human capabilities in scientific research. Termed Artificial SuperIntelligence Robots (ASIR), these systems operate entirely independently across all environments—virtual, physical, and experimental—and are capable of conducting groundbreaking research without any human intervention. They not only synthesize knowledge across disciplines but also innovate and formulate entirely new scientific principles. Their work leads to unprecedented scientific discoveries, positioning them as pioneers at the forefront of AI-driven research. While acknowledging the inherent uncertainties in achieving Level 5 autonomy due to substantial technical, ethical, and practical challenges, this level serves as an ambitious long-term goal for the field, inspiring continued exploration and innovation in autonomous scientific discovery.

\begin{table}[ht]
\vskip 0.2in
\begin{center}
\centerline{\includegraphics[width=0.99\columnwidth]{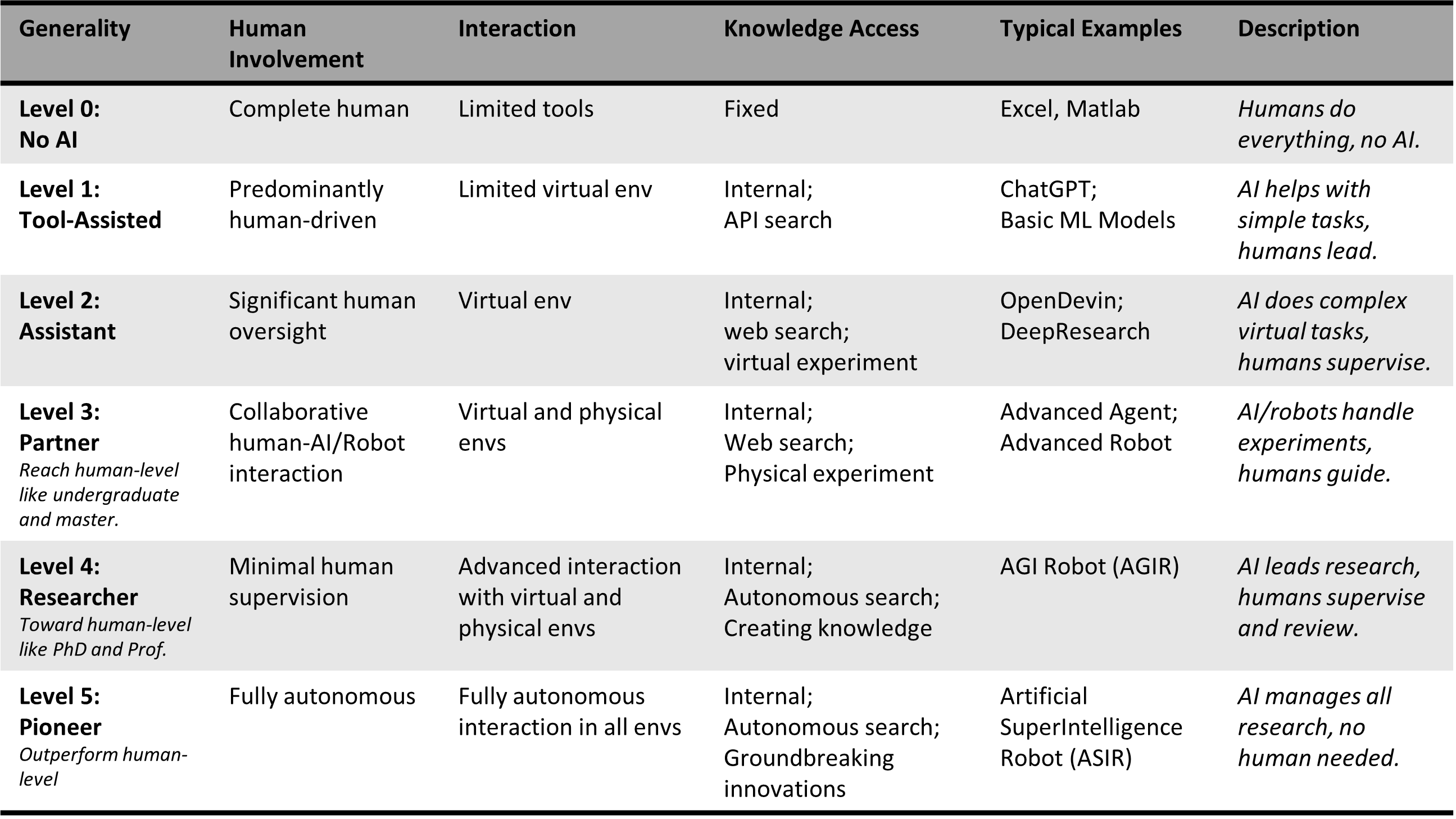}}
\caption{Levels of Autonomous Generalist Scientist.}
\label{tab:ags_levels}
\end{center}
\vskip -0.2in
\end{table}

\begin{figure}[ht]
\vskip 0.2in
\begin{center}
\centerline{\includegraphics[width=1.00\columnwidth]{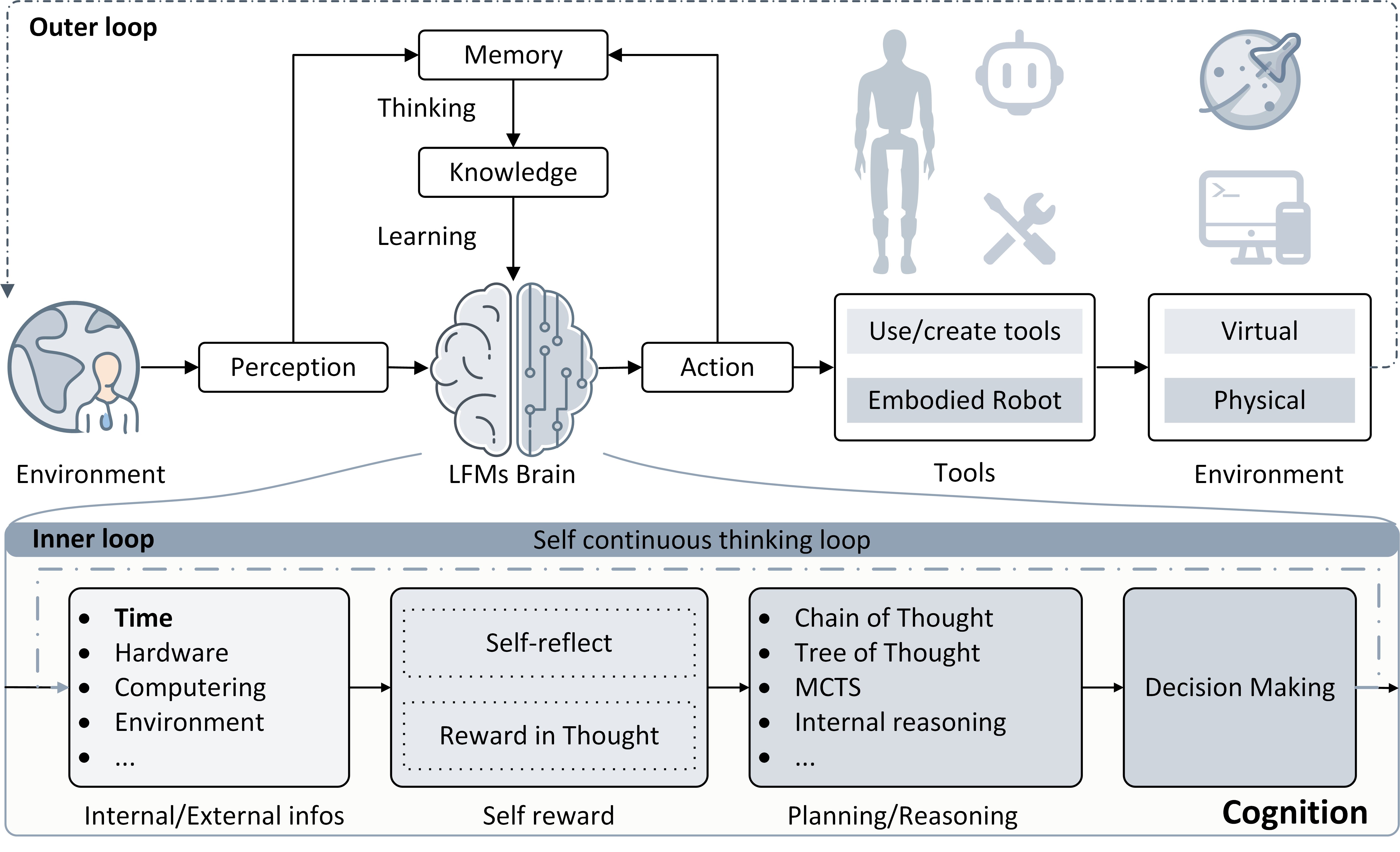}}
\caption{Framework of AGS brain.}
\label{ags_brain}
\end{center}
\vskip -0.2in
\end{figure}

\section*{Roadmap to Automatic Research with AI Scientist and Robot Scientist}

\subsection*{Overview}

The AGS offers a unified framework that blends cutting-edge AI with robotics to fully automate the research process (see Fig.\ref{ags_framework}, Fig.\ref{ags_brain}). Built on a multi-agent system, it pairs agentic AI and embodied robotic system with general purpose manipulation capabilities. The AI agents handles virtual tasks like coding, hypothesis creation, and data analysis, while robotics takes on physical duties, such as operating lab tools and running precise experiments. This combination speeds up research, improves accuracy, and ensures reproducible results, paving the way for a game-changing shift in multidisciplinary science.

\subsection*{Literature Review}

The literature review underpins research by pinpointing existing knowledge, gaps, and new possibilities. Traditionally, it relies on manual searches and analysis of countless papers—a slow process often limited by outdated data access ~\cite{wagner2022artificial}. This section explores the shift to AI-driven methods, contrasting conventional database or API approaches with advanced OS agent-driven systems that emulate human actions for complex searches and tasks in virtual settings.

\subsubsection*{Limitations in Traditional Review Approaches}

Conventional automated literature reviews lean on manual effort or restrictive database or API access, narrowing the scope and freshness of data. Database searches lag due to indexing delays, while API-based tools, though faster, they face significant limitations as many scholarly journals and publishers simply do not provide API access to their publications. Systems like Survey Agent ~\cite{wang2024surveyagent} and AutoSurveyGPT ~\cite{xiao2023autosurveygpt} use conversational AI and GPT models to speed up reviews, and specialized tools like the AI Chatbot in Cancer Research ~\cite{pan2023assessment} aid niche fields.  Despite improving upon manual methods, these API-driven systems remain constrained by data sources, critically limiting access to cutting-edge research in rapidly evolving fields and underscoring the need for more sophisticated approaches.

\subsubsection*{Autonomous and Comprehensive Information Acquisition by OS Agents}

To overcome these hurdles, OS agents mimic human-like interactions with digital platforms, moving beyond static API limitations by directly interfacing with websites and applications as human researchers would. Tools like GPT-4 Vision ~\cite{zheng2024gpt} leverage visual understanding to handle complex web tasks including accessing journal websites without APIs, interpreting search results, and extracting data from diverse publication formats. OS-Copilot ~\cite{wu2024copilot} advances this paradigm through continual self-improvement mechanisms that enable adaptation to changing digital interfaces and learning from past interactions—crucial capabilities when navigating the heterogeneous landscape of academic repositories. Multimodal agents have further extended these capabilities, with VisualWebArena ~\cite{koh2024visualwebarena} providing rigorous benchmarking across realistic literature search scenarios and OSWorld ~\cite{xie2024osworld} enabling sophisticated navigation through institutional authentication gates, publisher websites, and citation networks that typically resist API-based access. Unlike their predecessors, these systems can dynamically retrieve information from previously inaccessible sources, including subscription-based journals, preprint servers, and conference proceedings, thereby gathering comprehensive, current information that encompasses the full spectrum of scholarly communication and strengthening the foundation for subsequent research tasks.

\subsubsection*{Intelligent Processing, Synthesis, and Gap Identification}

Once the relevant literature is acquired, the AGS framework proceeds with intelligent processing and knowledge extraction using advanced reasoning models. This involves analyzing the content of the retrieved documents to identify key concepts, methodologies, findings, and conclusions within each publication. Moving beyond simple keyword extraction, the system aims to understand the semantic relationships between different pieces of information, identify the main arguments and evidence presented, and extract structured data where possible. The processed information is then subjected to a synthesis and pattern recognition phase. The framework analyzes the extracted knowledge to identify overarching themes, recurring methodologies, and significant trends across the literature. More importantly, it focuses on pinpointing gaps in the current understanding, inconsistencies in findings, and areas where further research is needed. By autonomously performing these sophisticated steps, the system establishes a strong foundation for guiding its subsequent scientific endeavors.

\autoref{tab:literature_review_method_comparison} illustrates three distinct approaches to automated literature review: knowledge-base systems relying on existing databases, search API-driven methods for web queries, and OS agents mimicking human-like interactions across digital platforms. OS agents offer significant advantages through their computer-using capabilities—visually interpreting interfaces, executing mouse and keyboard operations, navigating authentication barriers, and processing diverse file formats without predefined APIs. These agents can traverse subscription paywalls via institutional credentials, extract data from interactive visualizations, follow citation networks across disparate platforms, and even manipulate search parameters to overcome indexed content limitations. Unlike previous methods, they can dynamically adapt to changing web interfaces and publisher policies, accessing the most current research regardless of structured data availability. This evolution from database-dependent methods to agent-driven approaches represents a paradigm shift in scientific inquiry, transforming literature review from a preliminary bottleneck into a dynamic, ongoing component of the research process. These advancements lay the foundation for fully automated scientific workflows where literature discovery seamlessly integrates with hypothesis generation and experimental design, accelerating the pace of innovation across disciplines.

\autoref{tab:literature_review_method_comparison} illustrates a clear progression in automated literature review methodologies, moving from static knowledge bases to API-driven queries and culminating in the sophisticated capabilities of OS agents that emulate human-like interaction across digital platforms. The unique computer-using abilities of OS agents—including visual interface interpretation, execution of user-like commands, navigation of authentication barriers, and versatile file format processing—provide significant advantages. Their capacity to access research behind subscription paywalls, extract data from dynamic visualizations, and traverse complex citation networks, coupled with their adaptability to evolving web interfaces and publisher policies, marks a fundamental paradigm shift in scientific inquiry. This evolution transforms the literature review from a traditionally static and potentially limiting initial step into a dynamic and continuously updated component of the research process. These advancements not only overcome previous bottlenecks in accessing comprehensive and current scientific knowledge but also lay a critical foundation for realizing fully autonomous scientific workflows, where intelligent literature discovery becomes an integral and ongoing driver of hypothesis generation, experimental design, and ultimately, the accelerated pace of innovation across all scientific disciplines.

\begin{table}[!ht]
\vskip 0.2in
\begin{center}
\centerline{\includegraphics[width=0.99\columnwidth]{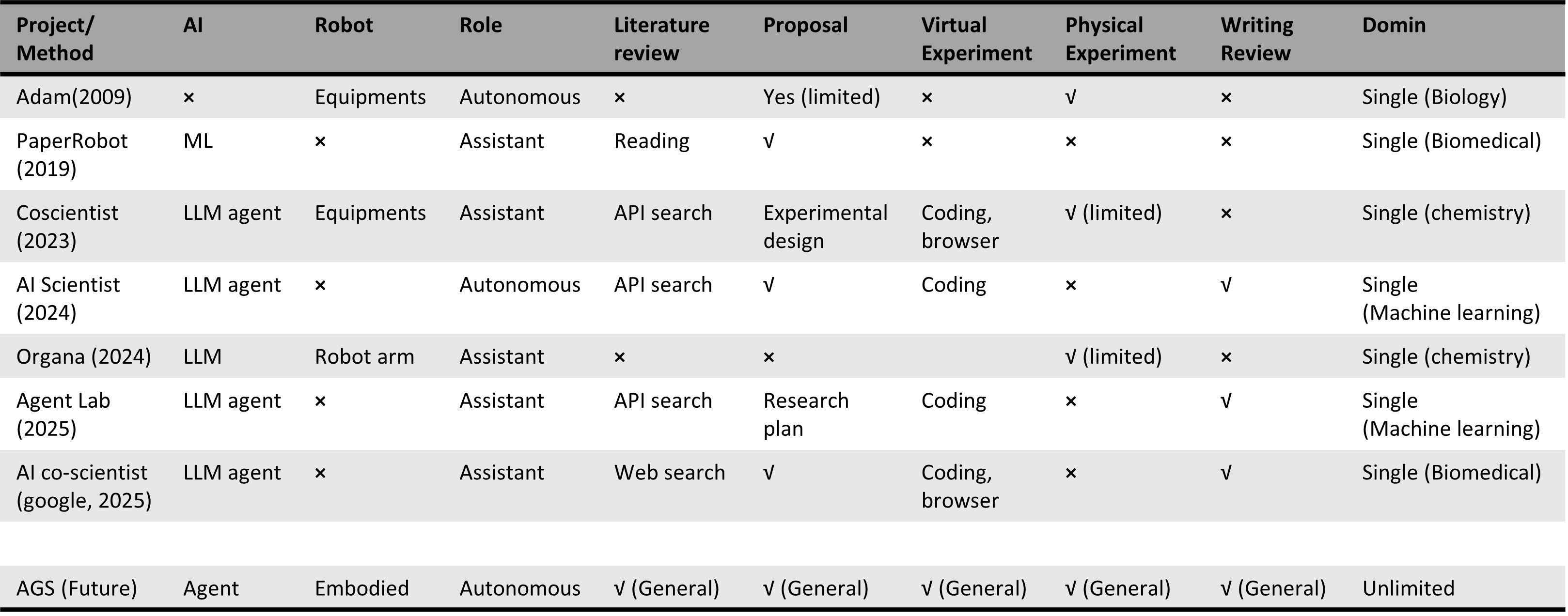}}
\caption{Comparison of current AI Scientists and Robot Scientists.}
\label{tab:ai_robot_scientist_method_comparison}
\end{center}
\vskip -0.2in
\end{table}

\begin{table}[!ht]
\vskip 0.2in
\begin{center}
\centerline{\includegraphics[width=0.99\columnwidth]{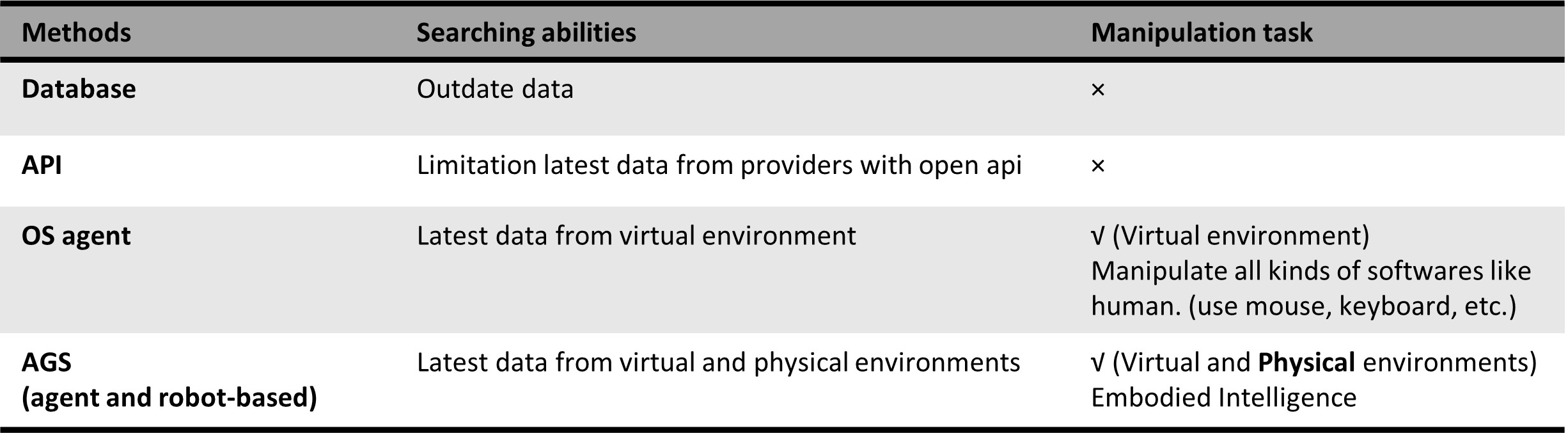}}
\caption{Comparison of methods for searching and performing manipulation tasks.}
\label{tab:literature_review_method_comparison}
\end{center}
\vskip -0.2in
\end{table}

\subsection*{Proposal Generation}

A research proposal maps out a study, pinpointing the problem and detailing a plan to tackle it. In NLP, LLM-generated ideas consistently demonstrate higher novelty than those produced by human experts ~\cite{si2024llmideas}. While other systems, such as those described in ~\cite{baek2024researchagent}, focus primarily on generating research ideas, the autonomous generalist scientist framework extends this capability through a comprehensive proposal development pipeline. This process begins with automated gap analysis across literature, identifying contradictions and unexplored connections. The system then proceeds through a structured workflow: formulating precise problem statements with clear research boundaries; generating testable hypotheses based on theoretical foundations; designing rigorous methodologies with appropriate controls and statistical considerations; and creating detailed implementation plans including timelines and resource requirements. Throughout this process, the AGS employs a multi-agent architecture where specialized components evaluate methodological soundness, novelty assessment, and feasibility analysis. The system could iteratively refines each proposal component through internal critique cycles and external feedback integration, ensuring proposals are both innovative and practically executable within the identified research landscape.

\subsubsection*{Problem Statement}

Crafting a research proposal begins with formulating a precise problem statement that defines the investigation's scope and significance. The AI system systematically analyzes literature review outputs through semantic relationship mapping and citation network analysis to identify knowledge gaps, contradictory findings, and emerging research frontiers. It employs bibliometric analysis to quantify research density across subfields, highlighting underdeveloped areas with high potential impact. The system then synthesizes these insights to formulate problem statements that balance specificity with broader theoretical relevance, ensuring research questions are both novel and anchored in established frameworks. Through hierarchical topic modeling and ontological clustering techniques, the AI transforms broad research domains into operationalizable inquiries with clear boundaries and testable components. Each candidate problem statement undergoes rigorous evaluation against criteria including theoretical contribution, methodological feasibility, and alignment with current scientific discourse, ensuring the resulting research direction is positioned to make meaningful advances while remaining tractable within practical constraints.

\subsubsection*{Hypothesis and Methodology}

Following problem definition, the AI system generates hypotheses through a systematic approach that evaluates potential research concepts against the existing corpus. It employs advanced computational strategies to assess candidate hypotheses, measuring their novelty against published findings and identifying conceptual intersections that remain unexplored. The system prioritizes hypotheses that bridge disciplinary boundaries or challenge established paradigms while maintaining theoretical coherence. For each promising hypothesis, the AI develops comprehensive testing methodologies using decision frameworks that evaluate various experimental designs against validity criteria. This includes selecting appropriate research approaches based on hypothesis structure and variable relationships. The system implements quantitative assessment techniques to determine optimal research parameters and integrates checks for potential confounds and bias sources. Drawing from its literature database, the AI incorporates methodological refinements from similar studies, adapting proven techniques and measurement protocols with documented reliability. This systematic approach ensures proposed methodologies maintain scientific rigor while remaining operationally feasible, establishing a solid foundation for empirical investigation that balances innovation with methodological soundness.

\subsubsection*{Research Planning}

The research planning phase transforms hypotheses and methodological designs into executable scientific workflows. The AGS system seamlessly transitions from conceptual formulation to operational planning by leveraging its comprehensive literature analysis to inform implementation strategies. It integrates insights from prior scientific workflows to structure research execution, extracting proven methodological frameworks while conducting multi-dimensional risk assessment across computational, experimental, and logistical domains. The planning module constructs a comprehensive timeline with phase-dependent resource allocation, establishing clear milestones for literature benchmarking, method validation, data collection, analysis, and publication preparation. It evaluates operational feasibility by calculating resource requirements against availability, identifying potential bottlenecks in experimental procedures, computational demands, and collaborative dependencies. For laboratory-based research, the system incorporates equipment calibration periods, material procurement timelines, and specialized personnel availability. For computational studies, it schedules processing time, storage requirements, and code validation phases. The system employs sensitivity analysis to identify critical path components, establishing contingency buffers at strategic intervals and decision gates where research direction may require adaptation. Through simulation of various research trajectories and their cascading effects on subsequent phases, the system enables dynamic project management that maintains momentum while remaining responsive to emerging findings, technical challenges, or unexpected opportunities that arise during implementation.

\subsubsection*{Iterative Refinement by Communication, Feedbacks}

The advanced research system could employ a sophisticated discourse architecture for proposal refinement, establishing bidirectional communication channels with domain experts, institutional stakeholders, and specialized evaluation agents. This framework facilitates the presentation of preliminary proposals through structured academic formats, complete with hypothesis articulation, methodological justification, and anticipated significance metrics. Upon dissemination, the system implements a systematic feedback collection protocol, parsing critiques through natural language processing to identify conceptual weaknesses, methodological limitations, and potential theoretical inconsistencies. The multi-agent peer review mechanism employs specialized evaluation modules—each calibrated to assess different proposal aspects including theoretical grounding, methodological rigor, statistical validity, and ethical considerations—creating a comprehensive critique landscape that mimics rigorous academic peer review. Through adaptive belief revision strategies, the system dynamically adjusts confidence weights for proposal components based on expert consensus or disagreement patterns, prioritizing revisions accordingly. This recursive refinement process continues through multiple iterations until convergence criteria are satisfied, with each cycle enhancing proposal coherence, methodological defensibility, and theoretical contribution. The resulting research framework undergoes final harmonization to ensure internal consistency across all sections, producing a submission-ready proposal that has effectively undergone pre-submission review scrutiny comparable to formal academic evaluation processes.

\subsubsection*{Innovation and Research Gap Alignment}

Moreover, the advanced research system needs implement a structured evaluation framework to quantitatively assess proposal innovation across multiple dimensions. Through comparative analysis against the contemporary research landscape, the system calculates innovation indices that measure both incremental advances and paradigm-shifting potential. This assessment employs citation network projection techniques to forecast how the proposed research might influence future knowledge trajectories within the field. The system evaluates potential impact through multiple lenses: theoretical contribution (advancing conceptual frameworks), methodological innovation (introducing novel techniques or applications), and translational potential (bridging research domains or theoretical-practical divides). This systematic approach extends beyond merely identifying research gaps to quantifying the proposal's strategic positioning within evolving research frontiers and emerging disciplinary intersections. By linking innovation metrics to specific knowledge deficits identified during literature analysis, the system ensures research initiatives address substantive gaps rather than superficial ones, maximizing the probability of meaningful scientific advancement while minimizing effort duplication across the research ecosystem.

\subsection*{Experimentation}

Scientific research encompasses a dual landscape of virtual and physical manipulations, with both domains crucial for comprehensive scientific inquiry as illustrated in Table \ref{tab:vp4sr}. This duality manifests across all disciplines—from physics requiring both theoretical modeling and equipment operation to social sciences demanding both data analysis and field research. Traditional scientific methodology relies heavily on human expertise to navigate this complex terrain, particularly in designing and executing physical experiments—an approach that is inherently resource-intensive and often creates bottlenecks in the research pipeline. While artificial intelligence has revolutionized virtual experimentation through advanced simulation, optimization, and data analysis capabilities, the automation of physical experimentation remains significantly underdeveloped. Current AI systems for scientific discovery \cite{ma2024llm} and data-centric applications \cite{hong2024data} excel in computational environments but fail to translate this intelligence into physical laboratory settings. This stark capability gap creates a fundamental limitation in achieving truly autonomous scientific research. The disparity becomes particularly evident in fields like chemistry and materials science, where virtual modeling can predict molecular behaviors, but physical synthesis and characterization still require manual intervention. These constraints underscore the critical need for embodied intelligent systems—robots capable of executing complex physical manipulations with the precision, adaptability, and contextual awareness characteristic of human scientists. Current robotic platforms, while advancing rapidly, still struggle with generalization across experimental contexts, typically excelling only in narrowly defined tasks without the flexibility to adapt to diverse experimental protocols or unexpected situations \cite{yoshikawa2023large, ma2024survey}. Addressing this capability gap represents one of the most significant challenges in developing a truly autonomous generalist scientist system capable of seamlessly integrating both virtual and physical experimentation.

\begin{table}[!ht]
\vskip 0.2in
\begin{center}
\centerline{\includegraphics[width=0.99\columnwidth]{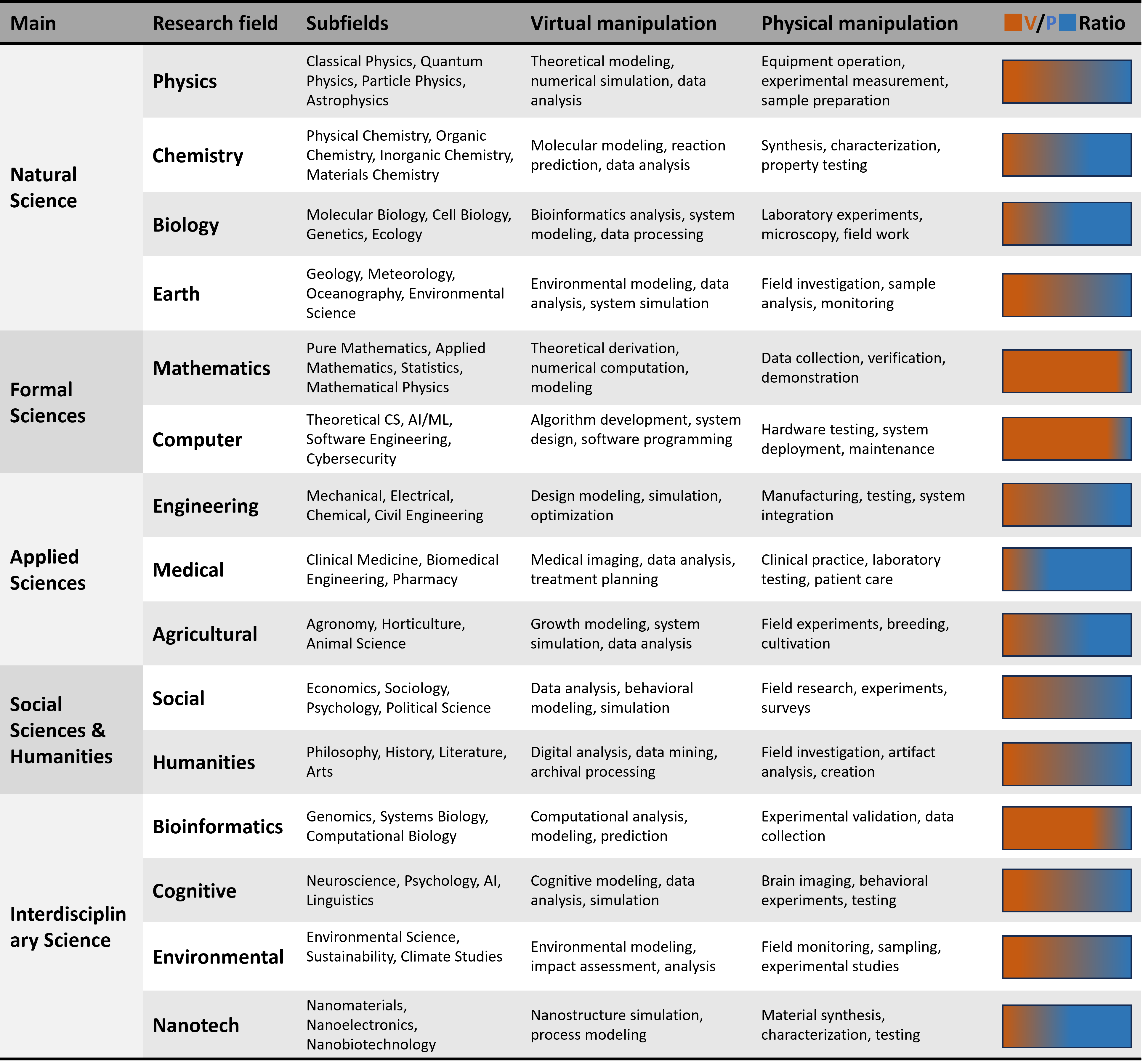}}
\caption{Virtual and Physical Manipulation Needs for Scientific Research. This table illustrates the characteristic requirements for virtual and physical manipulation across diverse scientific disciplines. The V/P ratio (rightmost column) represents general tendencies rather than precise quantitative measurements, highlighting the relative emphasis typically placed on computational versus experimental approaches in each field.}
\label{tab:vp4sr}
\end{center}
\vskip -0.2in
\end{table}

\begin{table}[!ht]
\vskip 0.2in
\begin{center}
\centerline{\includegraphics[width=0.99\columnwidth]{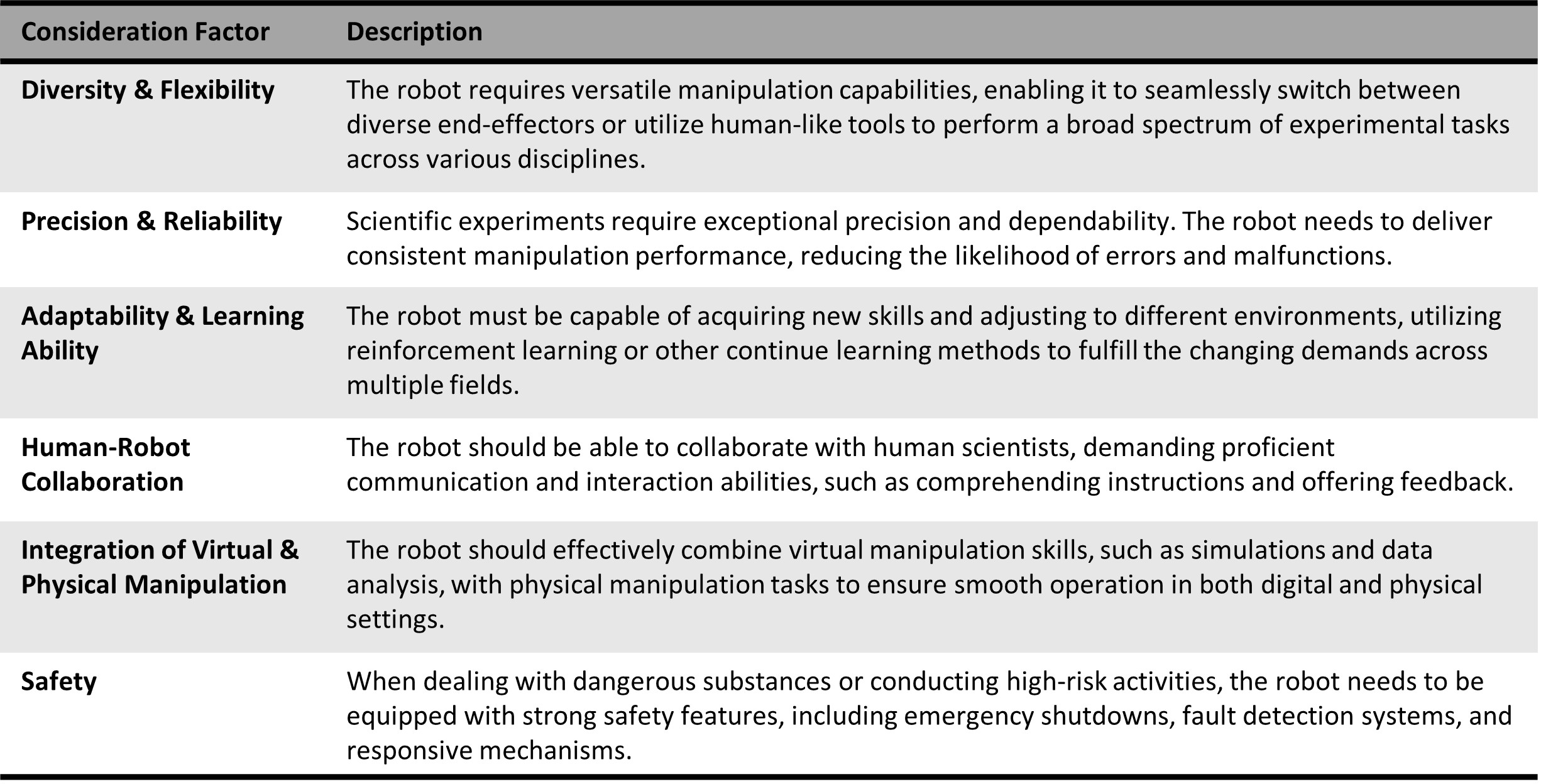}}
\caption{Comprehensive Considerations for Manipulation Needs.}
\label{tab:considerations}
\end{center}
\vskip -0.2in
\end{table}

\subsubsection*{Current Advances and Remaining Challenges in Experimentation}

Current Virtual Experimentation Capabilities of AI agent: Recent initiatives, such as AI Scientist \cite{lu2024ai}  and AI co-scientist \cite{gottweis2025towards} frameworks have demonstrated promising capabilities in automating specific aspects of the scientific research process, yet exhibit substantial limitations in their virtual manipulation competencies, not to mention their complete inability to conduct physical laboratory work \cite{castelvecchi2024researchers, gao2024empowering, ifargan2024autonomous}. These systems excel within narrowly defined computational domains—executing predetermined algorithms, performing parameter optimization, and conducting statistical analyses on structured datasets—but lack the comprehensive computer-using proficiencies that characterize human scientific practice. Human researchers fluidly transition between diverse computational environments throughout the research workflow, a versatility that current AI systems fundamentally cannot replicate. These platforms demonstrate significant deficiencies in navigating the complex landscape of scientific literature repositories, which often feature heterogeneous interfaces, authentication requirements, and organizational structures. They struggle to effectively utilize the specialized scientific software ecosystem, including computational modeling environments, analytical tools, and simulation frameworks that frequently demand nuanced configuration and cross-platform integration. Their capabilities in scientific visualization remain rudimentary, failing to generate the sophisticated graphical representations essential for data interpretation, hypothesis communication, and result dissemination across scientific disciplines. This limitation extends to the creation of publication-quality figures conforming to disciplinary conventions, the development of conceptual diagrams illuminating complex phenomena, and the production of interactive visualizations enabling dynamic data exploration. Perhaps most critically, current systems lack the metacognitive flexibility to identify appropriate computational tools for emerging research questions and to adaptively orchestrate workflows spanning multiple software environments as investigations evolve. This profound capability gap constitutes a significant barrier between algorithmic reasoning and practical scientific implementation, impeding progress toward autonomous end-to-end scientific discovery systems. The disparity between sophisticated reasoning capabilities and limited virtual manipulation competencies represents a fundamental constraint on the realization of fully automated open-ended scientific research platforms. Moreover, the complete absence of physical experimentation capabilities in these agent platforms fundamentally restricts their scientific scope to purely computational domains, excluding vast territories of empirical science that require direct interaction with physical phenomena. This limitation represents an even more formidable challenge that must be addressed to realize truly comprehensive autonomous scientific systems, as we will explore in the following section on physical experimentation capabilities.

Developments in Physical Experimentation of Robotic Systems: While virtual experimentation systems face significant limitations, parallel developments in robotic systems for physical experimentation have emerged across scientific domains. Specialized platforms for autonomous chemical research \cite{boiko2023autonomous, m2024augmenting} demonstrate notable progress in executing precise experimental protocols under controlled laboratory conditions. These systems can perform consistent material handling, precise measurements, and reproducible reaction procedures that reduce human error in experimental workflows. However, current robotic implementations remain fundamentally constrained by their domain-specificity and operational rigidity. Unlike human scientists who fluidly adapt experimental approaches based on unexpected observations, existing robotic platforms typically execute predetermined procedural sequences with minimal capacity for experimental improvisation or protocol adaptation. They operate effectively within narrowly defined parameter spaces but struggle when confronted with experimental anomalies, unexpected material behaviors, or equipment malfunctions that routinely challenge human researchers. Despite advances in robotic learning leveraging comprehensive datasets \cite{Openxvuong2023open, padalkar2023open}, current systems exhibit limited generalization capabilities across diverse experimental contexts. This profound limitation stems from their development as specialized instruments rather than versatile research partners—they excel at executing predefined experimental protocols but lack the universal manipulation skills, adaptive motion planning, and contextual awareness necessary for open-ended scientific discovery. Scientific experimentation demands exceptional dexterity across diverse physical interactions—from delicate micromanipulation of biological specimens to precise assembly of complex apparatus—capabilities that remain beyond current robotic systems. The significant gap between specialized robotic platforms and the versatile physical capabilities of human scientists highlights the critical need for general-purpose embodied AI robots equipped with both universal manipulation skills and generalized, flexible motion capabilities that can operate across experimental domains, adapt to unforeseen circumstances, and perform the diverse physical interactions that comprehensive scientific inquiry demands.

\subsubsection*{Advancing General-Purpose Robotic Systems with Embodied AI}

Employing LLMs within embodied AI frameworks presents a compelling approach to connect high-level cognitive processes with physical actions in real-world settings. Embodied AI systems feature agents designed to interact purposefully with their surroundings via perception, reasoning, and motor control. LLMs augment these systems by infusing them with advanced natural language processing and reasoning, empowering robots to interpret intricate instructions, assimilate knowledge from varied sources, and formulate context-sensitive decisions \cite{ma2024survey}. This integration enables robots to tackle a broader spectrum of tasks, adjusting to evolving objectives and circumstances, thereby moving beyond specialized functions toward greater versatility. Platforms engineered for generalist agents, like \textit{OpenDevin} \cite{wang2024opendevin, brohan2023rt}, illustrate potential future capabilities, though applying such systems effectively to physical sciences poses considerable difficulties. While vision-language models \cite{ma2024survey} and embodied AI research indicate a potential pathway for linking complex directives to real-world execution, the technology remains nascent concerning scientific experimentation. Publicly available large-scale robotic learning datasets \cite{Openxvuong2023open, padalkar2023open} foster transparency and interdisciplinary collaboration, potentially improving robot-assisted experiments; however, achieving robust generalizability and scalability in complex, dynamic environments continues to be a challenge.

Currently, research into world models concentrates on agents learning through environmental interaction. World models serve as vital components for general-purpose robots, enabling the construction of internal environmental representations and the prediction of action consequences. By learning spatial, temporal, and causal environmental relationships, these models permit robots to function within complex, unstructured settings. Robots utilizing world models can navigate and interact with objects in novel situations by simulating potential scenarios, forecasting action outcomes, and selecting optimal strategies informed by sensor data, machine learning, and probabilistic techniques. A well-developed world model crucially allows a robot to generalize from prior experiences to new contexts, a fundamental characteristic for achieving autonomy and adaptability. Coupled with embodied AI progress, world models aid in developing robots capable of flexible, intelligent decision-making across diverse real-world applications, moving beyond task-specificity  \cite{abou2024physically, wu2023daydreamer}.

Combining world models with LLM-based embodied AI marks a significant step forward for general-purpose robotics, as it unites structured environmental representations with sophisticated cognitive capabilities \cite{aljalbout2024limt}. LLMs offer sophisticated cognitive functions, allowing robots to process complex language and reason across scenarios. And, world models furnish a structured internal representation of the physical environment, facilitating outcome prediction and real-time adaptation. The synergy is critical: LLM-derived linguistic reasoning guides decision-making, while world models ground these decisions in the practical constraints and dynamics of the physical world. For instance, an LLM could help a robot understand a high-level command like "prepare the lab bench for the next experiment," while the world model ensures the robot can navigate the space, anticipate movement consequences, and adjust to unexpected environmental changes. This combination yields robots possessing enhanced intelligence and adaptability, capable of performing diverse tasks with greater autonomy and contextual understanding, effectively bridging abstract thought and physical action.

The primary obstacles facing general robotic systems in physical environments include:
\begin{itemize}

    \item \textbf{Robust Perception and Manipulation}. General-purpose robots require sophisticated environmental awareness and interaction capabilities \cite{xia2022review,zhang2025towards}. This encompasses accurate object recognition, spatial localization, and precise manipulation. Effective robotic systems depend on integrated sensor arrays and advanced actuator mechanisms that enable detailed environmental perception and fine-grained control precision.
    
    \item \textbf{Autonomy and Decision-Making}. Effective robotic systems must demonstrate independent reasoning and task execution capabilities \cite{wu2024safety}. This necessitates sophisticated planning algorithms, contextual reasoning frameworks, and adaptive learning mechanisms. Modern robots must navigate dynamic environments, identify and circumvent obstacles, and respond appropriately to changing operational conditions. Research initiatives like \cite{10517611} are advancing autonomous decision-making frameworks that enable robots to independently plan and execute complex task sequences.
    
    \item \textbf{Adaptability and Generalization}. A key challenge is that robots could transfer knowledge between domains and apply previous learning to unfamiliar scenarios \cite{ju2024robo}. This requires sophisticated learning architectures capable of cross-domain knowledge application. Truly versatile robotic platforms must demonstrate flexibility across diverse operational environments and task requirements. Contemporary research such as \cite{bharadhwaj2024roboagent} focuses on developing learning frameworks that maximize generalization from limited training examples and enhance adaptation to novel contexts.
    
    \item \textbf{Physical Safety}. Human-robot collaboration introduces potential safety concerns, particularly in unstructured environments \cite{wu2024safety}. Ensuring robots operate safely while manipulating objects remains a critical priority. Research initiatives like \cite{10517611} emphasize developing safety-oriented behaviors through real-time environmental sensing and risk-aware learning models. Robots operating in shared spaces must make rapid safety assessments during task execution. Advanced systems including DeepMind's AutoRT \cite{ahn2024autort} implement comprehensive safety protocols, such as force limitation mechanisms and human-proximity operational constraints. The SafeVLA framework \cite{zhang2025safevla} integrates safety considerations into vision-language architectures to protect environmental elements, hardware systems, and human collaborators.
    
    \item \textbf{Human-level interaction}. Creating natural robot-human communication remains technically challenging \cite{ajoudani2018progress}, requiring advanced natural language processing \cite{ren2023robots}, emotional recognition capabilities, and non-verbal communication understanding. Robots must adapt to established social conventions and interaction protocols. Successful embodied AI depends on seamless human-robot engagement. This includes interpreting emotional states, understanding physical gestures, and recognizing social dynamics—all representing active research challenges.
    
    \item \textbf{Ethical and Legal}. Increasing robot autonomy raises significant ethical questions regarding decision processes and potential harm risks \cite{valluri2024exploring}. Critical considerations include responsibility allocation, privacy protection, and ethical data utilization. Robots interacting with humans must demonstrate sound ethical and moral reasoning. This becomes particularly significant in sensitive contexts like healthcare and eldercare where human wellbeing is directly impacted \cite{elendu2023ethical}.
    
\end{itemize}

\subsubsection*{Integrating Agentic AI and Embodied Robotics in Scientific Experimentation}

The integrated AGS artificial intelligence with advanced robotics to streamline experimental processes across virtual or digital experiments and laboratory environments. Drawing upon contemporary innovations, including language model applications for experimental parameter optimization \cite{ma2024llm} and breakthroughs in precision robotic handling of research materials \cite{li2024chemistry3d}, this comprehensive framework aims to enhances experimental adaptability and efficiency and develop a highly versatile and resource-efficient approach to scientific investigation.

\begin{itemize}
    \item \textbf{Experimentation Planning}: The system begins with interpreting research proposals, identifying the necessary experimental tasks, and creating a detailed execution plan. This involves not only the selection of appropriate tools and methods but also the management of resources and scheduling of tasks to ensure efficiency and accuracy. To achieve this,  reasoning methods like Chain of Thought (CoT)\cite{wei2022chain} will be integrated, enabling the system to decompose complex tasks into sequential, manageable steps, ensuring logical consistency and adaptability throughout the experimental process.

    \item \textbf{AI Agents for Virtual Experimentation}: Agentic AI plays a pivotal role in automating and enhancing virtual experimentation. These intelligent agents possess the capacity to fully interact with computer systems, enabling them to execute algorithms, conduct sophisticated data analysis, and process logical and textual information \cite{lu2024ai}. This empowers them to perform a wide range of computational experiments in domains such as machine learning, bioinformatics, mathematics, and AI for Science, etc. Furthermore, these agents are adept at designing and executing complex simulations \cite{jablonka202314}, providing invaluable insights and predictions that inform subsequent physical experiments.
    
    \item \textbf{Robotics for Physical Experiments}: Physical experimentation remains a cornerstone of scientific inquiry across virtually all disciplines (Table \ref{tab:vp4sr}). The framework leverages embodied intelligent robots to execute complex physical manipulations with precision and adaptability. Drawing upon advancements in flexible automation, these general-purpose robots are capable of performing diverse experimental protocols, handling materials, operating equipment, and making real-time adjustments based on sensory feedback. This capability addresses the current limitations of manual experimentation, enhancing efficiency and reducing human error \cite{darvish2024organa}.

    \item \textbf{Resource Management and Real-Time Adjustments}: Efficient allocation and management of experimental resources, including reagents, devices, and equipment time, and crucially, internal resources such as computational power, power, body status and operational duration, is paramount for research productivity. The framework incorporates mechanisms for dynamic resource management and the ability to adapt experimental protocols in real-time based on incoming data and intermediate results. Integrating LLMs with robotic systems, as demonstrated by platforms like \textit{ROS-LLM} \cite{mower2024ros}, facilitates structured reasoning and informed decision-making during the experimental process, optimizing resource utilization and experimental outcomes.
    
    \item \textbf{Ensuring Reproducibility and Accuracy}: The framework prioritizes experimental reproducibility and accuracy as scientific cornerstones. By employing validated robotic systems capable of precise execution and AI models adaptable to varying experimental conditions, this framework aims to enhances the consistency and reliability of experimental results. This approach mirrors the robust protocols seen in initiatives like \textit{Chemistry3D} \cite{li2024chemistry3d,10.7554/eLife.67995,Camerer2018EvaluatingTR}, aiming to establish a new standard for experimental rigor across diverse scientific inquiries.
    
\end{itemize}

\begin{table}[!ht]
\vskip 0.2in
\begin{center}
\centerline{\includegraphics[width=0.99\columnwidth]{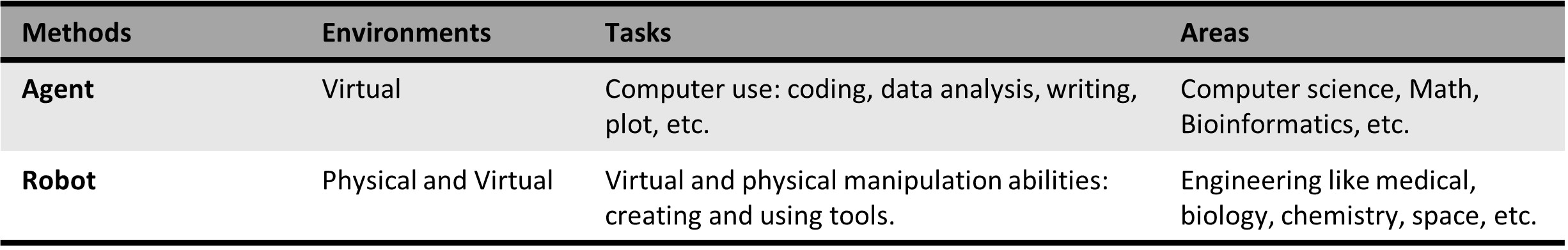}}
\caption{Comparison of agent and robot methods for research tasks.}
\label{tab:methods_comparison}
\end{center}
\vskip -0.2in
\end{table}

\subsection*{Manuscript Preparation}

The documentation and presentation of research findings represents a crucial component in the scientific workflow. This stage involves synthesizing experimental outcomes, organizing information logically, and communicating discoveries effectively to the academic community. Researchers traditionally face numerous challenges during this process, including ensuring factual precision, complying with disciplinary conventions, and articulating complex concepts in accessible language.

\subsubsection*{Automated Manuscript Drafting}

For manuscript drafting process, we propose utilizing state-of-the-art AI systems to bridge the gap between experimental results and scholarly documentation. This approach aims to create intelligent systems capable of producing preliminary manuscript drafts that organize research findings into structured sections following established academic conventions.

\begin{itemize}

    \item \textbf{Experiment Result Data Analysis and Summary}: The proposed system commences manuscript preparation by empowering AI agent to autonomously analyze the experimental outcomes derived from both virtual and physical investigations. This agent employs a diverse array of analytical methodologies, leveraging its ability to utilize existing scientific software, execute pre-defined algorithms, and even generate custom code to extract meaningful insights from the raw data. The agent's analytical process includes identifying key trends, performing statistical analyses, and generating concise summaries of the findings. To facilitate understanding and initial interpretation, the AI agent will also perform preliminary data visualization, selecting appropriate chart types to represent the core results \cite{yang2024matplotagent, vazquez2024llms}. This initial visualization serves as a crucial step in the data analysis pipeline, enabling the agent to identify significant patterns and prepare the ground for more sophisticated visual presentation in subsequent stages of manuscript preparation. This autonomous analytical capability underscores the system's ability to not just collect data, but to actively process and interpret it, demonstrating a significant step towards automated scientific reasoning.

    \item \textbf{Diversified Content Integration}: The framework incorporates sophisticated capabilities for producing a diverse array of scientific content formats, like figures and videos \cite{xing2024empowering, zhou2024survey}, crucial for conveying the multifaceted nature of research findings. Beyond traditional text, the system can generate quantitative tables with statistical rigor, insightful analytical graphs optimized for data interpretation, detailed procedural illustrations for complex experimental setups, and dynamic visual demonstrations to elucidate key processes or phenomena. Leveraging multiple representational approaches, the system intelligently selects the most effective visualization techniques to create professional-grade figures suitable for publication. Furthermore, for intricate methodologies, the framework can generate clear procedural demonstrations, potentially including animated sequences or interactive simulations. When appropriate, it can also develop interactive visual tools that allow readers to explore complex datasets or models directly. This comprehensive and adaptable approach to content integration ensures a richer and more accessible presentation of research outcomes across complementary formats, catering to diverse learning styles and enhancing the overall impact of the scientific communication.

    \item \textbf{Citation Coordination}: Efficient and accurate management of citations and references is a cornerstone of scholarly integrity \cite{bom2023exploring}. The system should excels in this critical task by seamlessly integrating with established or self-defined reference management software (e.g., Zotero, Mendeley, EndNote, as well as file-based formats like BibTeX which common in LaTeX, and etc.). This integration allows the AI to automatically ensure that all in-text citations are correctly formatted according to the target journal's style guidelines, eliminating a significant source of error and time investment for researchers. Simultaneously, the system maintains a comprehensive and up-to-date bibliography, verifying the accuracy and completeness of all cited works throughout the documentation process. This meticulous approach to citation coordination not only enhances the professionalism of the manuscript but also strengthens its credibility and facilitates the verification of sources by the scientific community.

    \item \textbf{Documentation Support}: Following the rigorous analysis of experimental data, the AI plays a pivotal role in facilitating the drafting of the manuscript. The system is capable of generating well-structured and coherent text for various essential sections, including the introductory context that establishes the research question and its significance, a detailed description of the methodological procedures employed, a clear presentation of the empirical outcomes, and an insightful interpretative discussion of the findings in relation to existing literature. Furthermore, the system provides intelligent assistance in the accurate representation of complex mathematical formulas, ensuring their correct syntax and formatting. By adhering to predefined manuscript templates specific to different journals or publication venues, the AI promotes consistency in structure and style, ultimately enabling researchers to focus their expertise on the core scientific content and narrative of their work, while minimizing the burden of formatting and structural conventions.
    
\end{itemize}

\subsubsection*{Peer Review Simulation}

To strengthen manuscripts before submission, the framework incorporates comprehensive evaluation protocols to identify and address potential weaknesses, ensuring adherence to scholarly standards.

\begin{itemize}

   \item \textbf{Internal Review Mechanisms}: The system employs dual evaluation approaches through reflexive assessment and collaborative agent critique. Its reflexive components evaluate argumentative structure, methodological robustness, and expositional clarity. Concurrently, specialized evaluation agents scrutinize distinct manuscript elements—including statistical methodology, experimental protocols, and theoretical frameworks—offering multifaceted improvement perspectives \cite{lingard2023writing}.
   
   \item \textbf{External Peer Review}: The framework facilitates engagement with external evaluators, including both AI systems and human specialists, to secure objective manuscript evaluation. These external review mechanisms simulate journal evaluation procedures, providing comprehensive feedback on scientific contribution, originality, and research significance. The system additionally facilitates human expert collaboration, integrating specialized knowledge to enhance document quality \cite{dergaa2023human}.
   
   \item \textbf{Ethical Considerations}: AI integration in scientific documentation raises important considerations regarding attribution and content authenticity. While language models demonstrate significant writing assistance capabilities, they present potential risks of generating inaccurate information or "hallucinations" \cite{liu2023overview}. The proposed framework incorporates governance protocols ensuring human researchers maintain appropriate oversight and responsibility for content development, preserving scholarly integrity throughout the documentation process.
    
\end{itemize}

\subsubsection*{Finalization and Submission}

The culminating phase involves comprehensive review, formatting refinement, and submission coordination. The framework streamlines these final processes to facilitate efficient manuscript publication.

\begin{itemize}

    \item \textbf{Journal-Specific Formatting}: The system implements precise formatting protocols according to target publication guidelines. This ensures all document elements—from typographical specifications to visual content placement—conform to journal requirements. The system applies appropriate reference styles, section organization, and visual presentation standards to meet publication criteria.

    \item \textbf{Submission Process Management}: The framework facilitates publication submission workflows. It manages submission documentation, coordinates file transfers, and processes editorial communications including revision requests. This automated approach streamlines interactions with publishing platforms while maintaining document integrity throughout the submission process.
    
\end{itemize}

The integration of AI into scientific documentation offers a transformative approach aimed at accelerating publication timelines while upholding rigorous quality standards, thereby enhancing accessibility to academic publishing across diverse research communities. This automation framework optimizes the temporal aspects of manuscript development, significantly reducing the cycle from initial drafting through meticulous refinement to final submission. This efficiency allows researchers to redirect valuable time and cognitive resources towards core scientific activities and intellectual advancement. Furthermore, by employing AI for critical tasks such as reference management, document formatting, and the generation of insightful visualizations, the system ensures a higher degree of technical consistency and accuracy, aligning manuscripts with established academic conventions and bolstering their reliability and professional presentation. Crucially, these automation capabilities have the potential to democratize access to professional-level document preparation, particularly benefiting researchers in resource-constrained environments or those with limited editorial support, ultimately fostering broader participation and a more equitable landscape within academic publishing.

Despite these considerable advantages, the realization of autonomous manuscript preparation presents ongoing challenges and necessitates focused future development. A primary limitation lies in the nuanced interpretation of complex data and the generation of truly novel insights, areas where deep domain expertise and creative reasoning remain critical. Ethical considerations surrounding authorship, intellectual property, and the potential for AI-generated inaccuracies also require careful navigation and the establishment of clear guidelines. Future research will therefore need to concentrate on enhancing the AI's capacity for sophisticated reasoning and contextual understanding, developing robust mechanisms for human oversight and error correction, and expanding its adaptability across the diverse spectrum of scientific disciplines. Addressing these challenges will be pivotal in unlocking the full potential of automated manuscript preparation to revolutionize the dissemination of scientific knowledge.

\subsection*{Reflection and Feedback}

In the generalist scientist framework, comprehensive information exchange and analytical self-assessment serve as critical components ensuring cohesive research progression. These mechanisms facilitate knowledge transfer between research phases, similar to collaborative dynamics in human research teams. Strategic module interaction coupled with systematic process evaluation enhances hypothesis formulation, methodological precision, and scientific output quality. This section examines how the AGS incorporates these interactive elements to maximize research effectiveness and innovation potential.

\subsubsection*{Internal Reflection}

A fundamental aspect of the integrated research automation architecture is its capacity for continuous monitoring and iterative enhancement of scientific processes, achieved through both analytical self-assessment and the seamless integration of insights across each research phases. Emulating the collaborative synergy inherent in human research teams, this mechanism facilitates performance analysis and strategic adjustments, ultimately strengthening subsequent investigative outcomes and the overall quality of scientific output.

\begin{itemize}

    \item \textbf{Analytical Assessment and Information Exchange}: Drawing from contemporary AI self-evaluation research, the framework implements comprehensive performance monitoring protocols across its interconnected functional components (literature analysis, proposal development, experimentation, documentation). This integrated approach encompasses output accuracy verification, data relevance examination, and research objective alignment analysis. The system establishes bidirectional information exchange pathways, creating comprehensive feedback networks where insights from one stage directly inform and refine others. For example, experimental outcomes inform documentation priorities and emphasis, while literature discoveries trigger proposal refinements. Through these systematic assessments and dynamic communication structures, the system proactively identifies and addresses potential inaccuracies and ensures research coherence through continuous information updates between specialized modules \cite{renze2024self, ji2023towards}.
    
    \item \textbf{Iterative Enhancement Mechanisms}: The framework employs cyclical and iterative refinement processes, systematically enhancing research hypotheses, methodological approaches, and scientific outputs based on emerging data and integrated self-evaluation. This structured approach ensures continuous improvement by incorporating insights from previous research cycles, adjusting computational processes, and progressively building upon accumulated knowledge to generate increasingly reliable and scientifically sound outcomes \cite{li2023metaagents, dong2024multi, wu2023autogen}.
    
\end{itemize}

\subsubsection*{External Insights}

The incorporation of diverse external viewpoints represents an essential research component, introducing alternative analytical frameworks and identifying potential enhancement opportunities.

\begin{itemize}

    \item \textbf{Human Supervision}: The system implements structured oversight mechanisms enabling researchers to monitor development and provide directional guidance. This human-augmented approach maintains alignment between system operations and investigator objectives while preserving scientific integrity. The collaborative interface allows researchers to shape the investigative process while leveraging computational capabilities.
    
    \item \textbf{Peer Review Simulation}: The framework incorporates publication assessment modeling that replicates scholarly review processes. This virtual evaluation generates constructive critiques that inform manuscript refinement prior to formal submission, addressing potential methodological or structural weaknesses  \cite{lin2023automated, yu2024automated, jin2024agentreview}. The simulation enables preemptive quality enhancement based on anticipated reviewer perspectives
    
\end{itemize}

While the proposed communication and reflection architectures demonstrate significant promise, several challenges remain in fully realizing their potential. These include the need for efficient coordination of intricate interactions between multiple modules, particularly when managing vast datasets and integrating diverse research disciplines. Furthermore, achieving optimal system performance requires a careful balance between autonomous processes and essential human oversight to safeguard the integrity and quality of the research. Future work will therefore focus on refining these communication and reflection mechanisms, enhancing their robustness and adaptability across a wider range of diverse and complex research scenarios.

\begin{figure}[H]
\vskip 0.2in
\begin{center}
\centerline{\includegraphics[width=0.99\columnwidth]{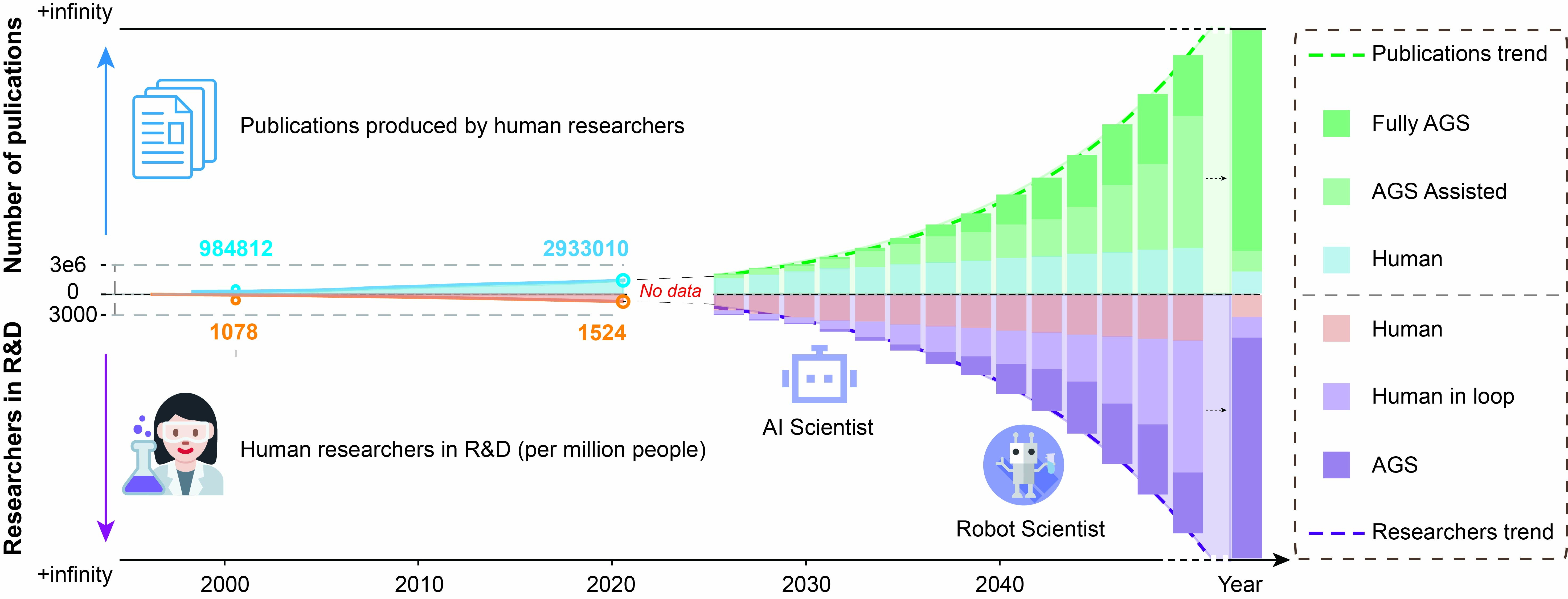}} 
\caption{Historical Patterns\cite{worldbank2024rd,worldbank2024-papers} and Projected Developments in Global Scientific Research Output and Workforce. (Note: Official World Bank Group data for the 2020-2024 period remains pending release.)}
\label{ags_trends}
\end{center}
\vskip -0.2in
\end{figure}
\section*{Open research questions}

\subsection*{How to Manage Publications from AI Scientists and Robot Scientists: Do We Need an Open Platform for Preprints?}  

\begin{figure}[ht]
\vskip 0.2in
\begin{center}
\centerline{\includegraphics[width=1.00\columnwidth]{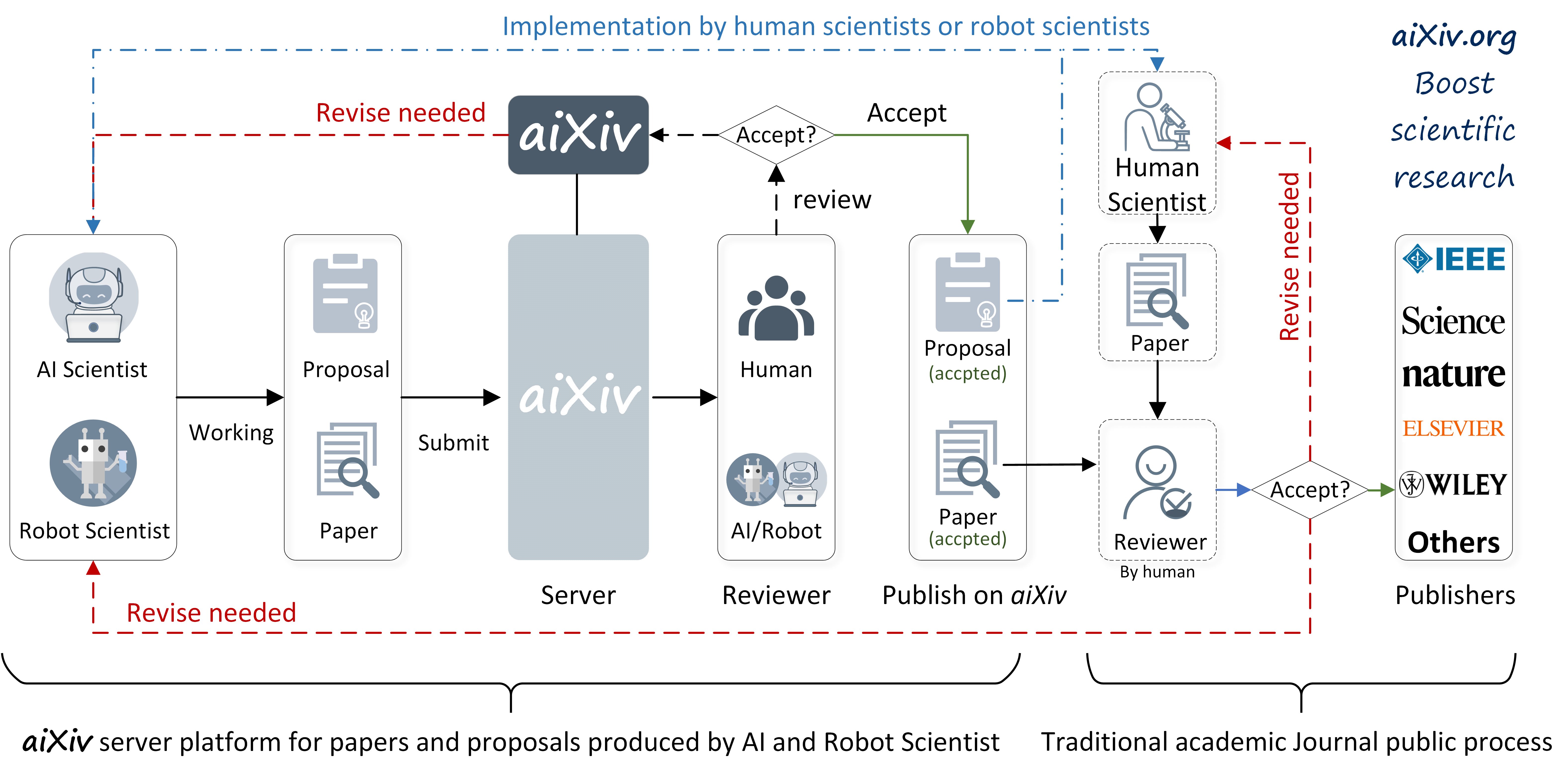}}
\caption{Framework of aiXiv server platform for papers and proposals produced by AI and Robot Scientist.}
\label{aixiv_framework}
\end{center}
\vskip -0.2in
\end{figure}

The emergence of AI scientists and robot scientist necessitates innovative approaches to manage and disseminate their research outputs. Recognizing that traditional academic systems, primarily designed for human researchers, may face challenges in handling publications and proposals generated autonomously, we propose the establishment of an intermediary platform, AIXIV (conceptualized in Fig.~\ref{aixiv_framework}), to bridge the gap between AI/Robot Scientists and the established academic publishing landscape. AIXIV would function as an open preprint server specifically for research generated by autonomous systems, implementing a tiered review process tailored to the unique characteristics of AI-driven discoveries. This approach aims to ensure that AI-generated research adheres to principles of transparency, credibility, and addresses ethical considerations pertinent to scientific communication involving non-human authors, while also facilitating their potential submission to traditional journals.

The AIXIV platform would function as a \textbf{public forum} where research outputs, in the form of both innovative proposals and comprehensive scholarly papers, generated autonomously by AI Scientists and Robot Scientists (representing non-human entities) can be submitted across a wide spectrum of scientific domains. As depicted in Fig.~\ref{aixiv_framework}, upon submission to the AIXIV server, these proposals and papers undergo a rigorous, multi-layered evaluation process. This review involves a combination of human experts and potentially AI or robot reviewers, leveraging the strengths of both forms of intelligence to assess submissions based on criteria such as feasibility, novelty, logical coherence, and the potential for significant scientific impact. Once a proposal is accepted and published on AIXIV, it can serve as a blueprint for further research, potentially implemented by human researchers or even by other AI or Robot Scientists, leading to subsequent paper submissions that would follow a similar review pathway. Furthermore, the AIXIV server would provide public Application Programming Interfaces (APIs) and user interfaces, facilitating easy access for both human and AI reviewers to examine submitted and published proposals and papers, thereby fostering a transparent and collaborative evaluation environment within the autonomous research community. Accepted proposals and papers are then published on the AIXIV platform (aixiv.org), providing immediate and open access to the research community. For completed research published on AIXIV, the platform aims to streamline the subsequent submission process to traditional academic journals, potentially boosting the visibility and impact of AI-driven scientific advancements.

While the AIXIV platform offers a promising pathway for integrating AI-generated research, several challenges must be addressed to ensure its successful adoption within the broader academic ecosystem. As depicted in Fig.~\ref{aixiv_framework}, clear guidelines for authorship attribution, accountability, and the validation of results originating from non-human agents will be crucial, both within AIXIV and upon potential submission to traditional journals. Human researchers may assume oversight roles to ensure the research meets established scientific standards. The platform will also need to tackle technical hurdles, including the development of unbiased evaluation metrics suitable for AI-generated content, the management of computational resources required for review processes, and the maintenance of system scalability. Moreover, ongoing dialogue with traditional publishers will be essential to determine their acceptance policies for papers originating from AIXIV's review process and to explore whether adapted evaluation criteria might be appropriate for distinguishing AI-generated from human-generated research. Despite these complexities, the establishment of a platform like AIXIV holds the potential to revolutionize scientific publication by fostering innovation, upholding academic integrity, and ultimately accelerating the pace of scientific discovery.

\subsection*{Does Robot Scientists need to be humanoid robots?}

The question of whether Robot Scientists should adopt a humanoid form is a nuanced one. While a humanoid design offers notable advantages, it is not strictly a prerequisite for effective function. The compatibility of humanoid robots with existing human-centric laboratory environments and research facilities is a significant benefit. Their inherent ability to navigate these spaces and utilize standard equipment without extensive modifications streamlines integration. Furthermore, advanced manipulation capabilities, particularly in the dexterous use of two hands, enable humanoid robots to perform intricate tasks, handle delicate instruments, and execute experiments demanding fine motor skills—abilities crucial across many scientific disciplines. The human-like form can also foster smoother and more intuitive interactions with human colleagues, potentially enhancing collaboration and communication within research teams. However, it is important to acknowledge that non-humanoid robotic designs, such as specialized mobile platforms equipped with advanced manipulators, can offer distinct advantages in specific scientific contexts. These designs may provide enhanced efficiency, precision for particular tasks, or the capacity to operate in environments unsuitable for humans. Ultimately, while humanoid robots offer compelling benefits for general-purpose scientific research and seamless integration into human-designed infrastructure, mobile robotic systems with sophisticated manipulation capabilities represent a viable and potentially more efficient alternative for a wide range of scientific investigations. The optimal form factor for a robot scientist will likely be determined by the specific research tasks and the environment in which it operates.

\subsection*{Can Robot Scientists Conduct Independent Scientific Inquiry in Extreme Environments Beyond Human Physical Limitations?}

Robot Scientists offer significant potential to extend scientific investigation beyond Earth's boundaries into space exploration (Fig.~\ref{ags_paradigm}), as well as into other extreme environments inaccessible or hazardous for humans. Initially establishing research capabilities on the Moon and Mars, these autonomous systems could methodically expand operations throughout our Solar System and potentially beyond. Similarly, Robot Scientists could revolutionize our understanding of Earth's deep oceans, operating under immense pressure, in perpetual darkness, and at vast distances from human support to explore hydrothermal vents, study unique ecosystems, and monitor geological activity. Furthermore, their capabilities extend to the micro and nano scales, where they could conduct independent scientific inquiry in areas such as advanced materials science, targeted drug delivery within the human body, or environmental monitoring of microscopic pollutants. Their operational advantages—functioning in harsh environments, making independent decisions despite communication delays (in space or the deep sea), and conducting continuous experiments—position them as invaluable assets for research in these challenging domains. Future development might enable robotic research networks operating across multiple celestial bodies, within the deepest ocean trenches, or even at the cellular level, facilitating detailed study of distant astronomical phenomena, unexplored marine ecosystems, and intricate microscopic processes. While formidable challenges exist in developing systems with sufficient environmental protection, long-term operational reliability, and effective distant communication protocols (where applicable), Robot Scientists represent a promising approach to expanding humanity's scientific reach and pushing the boundaries of knowledge across multiple frontiers. With continued technological advancement, these systems could substantially enhance our understanding of the cosmos, the Earth, and even the fundamental building blocks of matter through methodical exploration and discovery beyond human physical limitations.
\section*{Impact Statement}

This paper establishes a structured classification framework for AI-driven autonomous scientific systems. The proposed taxonomy facilitates precise communication between scientific researchers, technological innovators, and regulatory authorities. Through detailed categorization of autonomy levels in scientific investigation, this framework offers methodological guidance for multidisciplinary tool creation, enhances collaboration across scientific domains, and addresses critical ethical implications in autonomous research deployment. System development must adhere to established ethical guidelines, like the 23 Asilomar AI Principles, while maintaining compliance with applicable regulatory structures and international standards for advanced AI applications. This disciplined approach to classification supports the responsible evolution of autonomous scientific platforms that enhance research capabilities while incorporating appropriate safeguards against potential complications.
\section*{Conclusion}

The autonomous generalist scientist presents a groundbreaking framework harnessing the comprehensive interdisciplinary knowledge within foundation models, while synergizing AI agent capabilities with embodied robotic systems to automate scientific inquiry across digital and physical domains. This unified approach automates scientific inquiry across both digital and physical domains by enabling rapid data processing, hypothesis generation, and automated virtual experiments—coupled with advanced real-world research implementations that bridge computational simulations and laboratory experiments. By transcending conventional research boundaries, the methodology not only advances established disciplines but also paves the way for entirely new avenues of investigation. Furthermore, the reproducibility inherent in both computational and robotic platforms hints at new scaling laws for knowledge discovery, potentially elevating research productivity beyond traditional human-centered methods. As this integrated paradigm evolves, the synergy between artificial intelligence and robotics is set to transform academic research and drive innovations with substantial societal impact.

\citestyle{nature}
\bibliographystyle{templates/sn-basic}
\bibliography{main}

\paragraph{Acknowledgments}

We express our appreciation to Xiaoshuang Wang, Yinjun Jia, Xiang Hu, and Zhenzhong Lan for their insightful discussions and contributions that enriched this work. Special acknowledgment to Yajuan Shi for her constructive feedback and technical assistance with graphical enhancements.

\paragraph{Competing interests}

The authors declare no competing interests.

\paragraph{Author contributions}
All authors contributed to the conceptual development, textual composition, research formulation, supplied evaluative assessment, disscussions and comments.


\end{document}